\documentclass{article}

\usepackage[preprint]{neurips_2026}

\usepackage[utf8]{inputenc} 
\usepackage[T1]{fontenc}    
\usepackage{hyperref}       
\usepackage{url}            
\usepackage{booktabs}       
\usepackage{amsfonts}       
\usepackage{nicefrac}       
\usepackage{microtype}      
\usepackage{xcolor}         
\usepackage{algorithm}
\usepackage{algpseudocode}
\usepackage{amsmath, amssymb}
\usepackage{multirow}
\usepackage{subcaption}
\usepackage{multicol}
\usepackage{amsthm}
\usepackage{graphicx}
\usepackage{enumitem}

\usepackage{amsthm}
\newtheorem{lem}{Lemma}[section]

\title{Making LoRA Identifiable: Orthogonal Alignment for Continual Learning}

\author{%
  Pratik Rakesh Singh\thanks{These authors contributed equally to this work.}
  \quad Mohammadi Zaki\footnotemark[1] \quad Akash Saha\\
  \textbf{Aneesh Mukkamala} \quad \textbf{Pankaj Wasnik}\\
  Sony Research India\\
  \texttt{\{pratik.singh,pankaj.wasnik\}@sony.com}
}

\begin{document}

\maketitle

\begin{abstract}
There is growing interest in applying Parameter-Efficient Fine-Tuning (PEFT) to Continual Learning, as it enables faster training, mitigates catastrophic forgetting, and facilitates better adaptation to new tasks. Prior approaches have explored a range of strategies, including extending new LoRA branches for each incoming task, training under orthogonal loss objectives across all task-specific LoRA modules, and employing gating mechanisms to integrate new and existing LoRA modules. However, these methods incur memory costs that grow linearly with the number of tasks, limiting their scalability. More recent approaches investigate training a single LoRA by leveraging asymmetries in its parameterization; however, they often suffer from insufficient representational alignment across the LoRA parameter space. To address these limitations, we propose Stiefel Optimization and Aligned Rotation (SOAR), a two-stage framework for continual LoRA learning. First, we constrain the $\mathbf{A}$ matrix to lie on the Stiefel manifold, thereby reducing coordinate misalignment to the space of orthogonal rotation matrices. Second, we introduce an Orthogonal Procrustes alignment procedure for the $\mathbf{B}$ matrix to estimate the optimal rotation, coupled with a per-rank retention merging strategy that quantifies directional agreement between representations of new and previously learned tasks. We conduct extensive experiments across multiple models and benchmarks, demonstrating that SOAR achieves strong continual learning performance without incurring additional memory overhead.

\end{abstract}

\section{Introduction}

Large language models are increasingly deployed in settings where they must continuously acquire new capabilities without degrading previously learned knowledge \citep{, 10.1145/3735633, wang2024comprehensivesurveycontinuallearning}. This has driven growing interest in continual learning (CL) under parameter-efficient fine-tuning frameworks such as LoRA, where task-specific updates are represented as low-rank modifications to a frozen backbone \citep{wang2023orthogonalsubspacelearninglanguage, liang2025gatedintegrationlowrankadaptation, qiao2025mergeforgetsinglelora, liang2024inflora}. While maintaining a separate adapter per task is effective, it results in linear growth in memory and inference costs. This motivates the more challenging setting of \emph{single-adapter continual learning}, where all task updates must be consolidated into a single shared LoRA module \citep{qiao2025mergeforgetsinglelora, fa2026decomposingcomposingefficientvisionlanguage}.
A common strategy in this setting is to iteratively merge task-specific LoRA updates. Concretely, for a pretrained weight matrix $\mathbf{W}$, LoRA models task adaptation as $\mathbf{W}\leftarrow \mathbf{W} + \mathbf{BA}$. Prior work has observed that the two LoRA factors play different roles, with the input factor $\mathbf{A}\in \mathbb{R}^{r\times n}$ defining a low-dimensional subspace and the output factor $\mathbf{B}\in \mathbb{R}^{m\times r}$ capturing task-specific adaptations \citep{qiao2025mergeforgetsinglelora, hao2024floralowrankadapterssecretly, zhu2024asymmetry}. However, effectively leveraging this structure in continual learning remains challenging. In particular, existing methods operate on LoRA factors that are not expressed in a consistent coordinate system, making merging fundamentally unstable.

This instability arises from a geometric property of the LoRA parameterization. The factorization $\Delta \mathbf{W} = \mathbf{B}\mathbf{A}$ is non-identifiable: equivalent updates can differ by any \textit{arbitrary invertible transformation} in the rank space. In particular, for any $\mathbf{R}\in GL(r)$,
$\mathbf{BA}=(\mathbf{BR})(\mathbf{R}^{-1} \mathbf{A})$, leading to infinitely many equivalent factorizations. 

As a result, LoRA adapters learned across tasks may reside in \textit{misaligned} latent coordinate systems. Even if $\mathbf{A}$ and $\mathbf{B}$ are treated asymmetrically, merging them without resolving this ambiguity can introduce interference due to representational misalignment rather than true task conflict \citep{han2026w2tloraweightsknow, kumar2025latentspacefactorizationlora}.

In this work, we propose Stiefel Optimization and Aligned Rotation (\textbf{SOAR}), a principled framework for single-adapter continual LoRA learning that resolves these issues through a geometric treatment of the factorization. Our approach builds on the asymmetric roles of $\mathbf{A}$ and $\mathbf{B}$, but makes this structure well-posed for continual merging. First, we constrain the input factor $\mathbf{A}$ to lie on the Stiefel manifold,  ensuring $\mathbf{A}\mathbf{A}^\top=\mathbf{I}$. This removes the general linear invariance of the LoRA factorization, \textit{reducing} the ambiguity from arbitrary invertible transformations $\mathbf{R}\in GL(r)$ to orthogonal transformations $\mathbf{R}\in \mathcal{O}(r)$, thereby inducing a consistent and non-redundant coordinate system across tasks  (see Lemma~\ref{lemma:keylemma}).

This combination leads to a coherent consolidation strategy: $\mathbf{A}$ restricts the latent coordinate system to an orthonormal basis, reducing ambiguity to rotations, while $\mathbf{B}$ captures task-specific information that can then be explicitly aligned and merged via orthogonal transformations. Empirically, SOAR consistently outperforms existing single-adapter and merge-based baselines across multiple continual learning benchmarks, demonstrating that resolving the geometric inconsistency in LoRA parameterization is critical for mitigating forgetting.

While Orthogonal Procrustes alignment resolves the global rotational ambiguity between tasks, it operates at the full-matrix level and does not enforce consistent alignment across individual rank components. In practice, different directions of the low-rank decomposition can undergo varying degrees of change during task adaptation: some remain consistent and reusable, while others are substantially modified. As a result, a single global interpolation coefficient is insufficient to capture this heterogeneity. To address this, we introduce a rank-wise, forgetting-aware merging strategy that adaptively interpolates between aligned factors based on directional compatibility and relative change, enabling more fine-grained consolidation of shared and task-specific information.

With both corrections in place, rank-wise interpolation of $\mathbf{B}$ becomes well-defined, enabling directional agreement and magnitude drift to be measured in a consistent coordinate system. This, in turn, allows prior information to be selectively retained in directions that exhibit strong geometric compatibility and limited drift.
The following section formalizes each of these components and the design principles that connect them. We summarize our contributions below:

\begin{itemize}[leftmargin=*]
    \item We propose \textbf{SOAR}, a framework for single-adapter continual LoRA learning that enables knowledge consolidation at constant memory cost by explicitly addressing \textit{representational misalignment} across tasks.

    \item We show that LoRA adapters trained on different tasks can be expressed in inconsistent latent coordinate systems, making direct merging unstable. To address this, we constrain the input factor $\mathbf{A}$ to the Stiefel manifold, which enforces orthonormal rank directions and yields a stable, shared coordinate system across tasks.

    \item We introduce an \textit{alignment-aware merging strategy} for the output factor $\mathbf{B}$ that first resolves the remaining rotational ambiguity via Orthogonal Procrustes alignment, and then performs \textit{rank-wise, forgetting-aware interpolation} based on directional agreement and magnitude drift.

    \item We empirically validate SOAR across multiple continual learning benchmarks and model families, demonstrating consistent improvements over sequential and merge-based baselines while maintaining the efficiency of a single adapter.
\end{itemize}

\section{Related Work}

Continual learning has been studied extensively in the deep learning literature (see \cite{wang2024comprehensivesurveycontinuallearning}). In this work, we focus specifically on continual learning under parameter-efficient fine-tuning (PEFT), where task adaptation is performed through lightweight low-rank updates rather than full-model optimization.

\paragraph{Parameter-efficient continual learning.}
A growing body of work has explored continual learning with LoRA and related PEFT methods. O-LoRA \citep{wang2023orthogonalsubspacelearninglanguage} constrains new LoRA updates to remain orthogonal to previous ones in order to reduce interference. GainLoRA \citep{zhao2024mergingloraslikeplaying} learns task-specific scaling factors over shared LoRA components for efficient adaptation across tasks. InfLoRA \citep{liang2024inflora} mitigates task interference through similarity-aware routing and selective adaptation of LoRA modules. SAPT-LoRA \citep{zhao2024saptsharedattentionframework} introduces structured sparsity and parameter selection mechanisms to preserve important task-specific information during adaptation. Most closely related to our setting, SLAO \citep{qiao2025mergeforgetsinglelora} studies single-adapter continual learning by maintaining a shared low-rank subspace across tasks while enforcing orthogonality constraints on task-specific components.

Unlike prior methods, SOAR explicitly addresses the geometric non-identifiability of the LoRA factorization. Our approach stabilizes the latent coordinate system through Stiefel-constrained optimization and resolves residual rotational ambiguity via Procrustes-aligned consolidation.

\paragraph{Model merging and weight-space composition.}
Related ideas have also been explored in model merging, where independently trained models or LoRA adapters are combined directly in weight space. Unlike continual learning, these methods typically assume simultaneous access to all task-specific adapters and therefore avoid sequential forgetting by design.

LoRA-LEGO \citep{zhao2024mergingloraslikeplaying} treats LoRA adapters as composable modules for multi-task composition without joint training. KnOTS \citep{stoica2024modelmergingsvdtie} aligns task-specific LoRA subspaces prior to merging in order to reduce interference. Task Arithmetic \citep{ilharco2023editingmodelstaskarithmetic} represents tasks as parameter-space vectors that can be composed through addition and subtraction operations. TIES-Merging \citep{yadav2023tiesmergingresolvinginterferencemerging} resolves conflicting parameter updates through trimming and interference-aware merging. OPCM \citep{tang2025mergingmodelsflyretraining} proposes a sequential, training-free framework for incrementally composing task-specific models without retraining.

In contrast to these approaches, SOAR operates in the continual single-adapter setting, where past task-specific adapters are not retained. Our method therefore focuses on stable sequential consolidation under memory constraints, rather than offline merging of independently available task updates.

\section{Motivation}

\subsection{Continual learning with single LoRA}
\label{subsec:continual_lora_asymmetry}

In LoRA, a frozen weight matrix $\mathbf{W}_0 \in \mathbb{R}^{m \times n}$ is 
augmented with a low-rank residual:
\begin{equation}
    \mathbf{W} = \mathbf{W}_0 + \Delta \mathbf{W}, \qquad
    \Delta \mathbf{W} = \mathbf{B}\mathbf{A},
    \label{eq:lora_background}
\end{equation}
where $\mathbf{B}\in\mathbb{R}^{m\times r}$, $\mathbf{A}\in\mathbb{R}^{r\times n}$, 
and $r \ll \min(m,n)$. We consider a sequence of tasks $\{\mathcal{D}_t\}_{t=1}^{T}$, 
where fine-tuning on task $t$ yields task-specific parameters 
$(\mathbf{B}^{(t)}_{\mathrm{ft}}, \mathbf{A}^{(t)}_{\mathrm{ft}})$. Our goal is to 
maintain a single consolidated adapter $(\mathbf{B}^{(t)}_{\mathrm{merge}}, 
\mathbf{A}^{(t)}_{\mathrm{merge}})$ after each task, so that inference uses:
\begin{equation}
    \mathbf{W}^{(t)} = \mathbf{W}_0 + \mathbf{B}^{(t)}_{\mathrm{merge}}\,
    \mathbf{A}^{(t)}_{\mathrm{merge}},
    \label{eq:merged_lora_weight}
\end{equation}
Thus, the model stores only one consolidated LoRA update, keeping adapter memory fixed at $\mathcal{O}((m+n)r)$ rather than growing as $\mathcal{O}(T(m+n)r)$. The main challenge, therefore, shifts from storing past adapters to consolidating their information: the merged adapter must preserve useful knowledge from earlier tasks while adapting to the current one.

This consolidation should respect the asymmetric roles of the LoRA factors \citep{qiao2025mergeforgetsinglelora, zhu2024asymmetry}. Directly merging the full product $\mathbf{B}\mathbf{A}$ discards the two-factor structure that makes LoRA parameter-efficient and interpretable. Likewise, applying the same merge rule to $\mathbf{A}$ and $\mathbf{B}$ ignores the distinction between input-side subspace selection and output-side task adaptation. 

A natural approach to consolidation is to directly merge the fine-tuned factors 
$(\mathbf{B}^{(t)}_{\mathrm{ft}}, \mathbf{A}^{(t)}_{\mathrm{ft}})$ and 
$(\mathbf{B}^{(t-1)}_{\mathrm{ft}}, \mathbf{A}^{(t-1)}_{\mathrm{ft}})$ obtained from 
sequential fine-tuning. However, this is fundamentally ill-posed. Although both 
adapters are optimized using the same procedure, the LoRA factorization 
$\Delta\mathbf{W} = \mathbf{B}\mathbf{A}$ is non-identifiable: for any invertible 
$\mathbf{R} \in GL(r)$, the reparameterization 
$(\mathbf{B}, \mathbf{A}) \mapsto (\mathbf{B}\mathbf{R},\, \mathbf{R}^{-1}\mathbf{A})$ 
yields an identical weight update. Consequently, the optimizer has no mechanism to 
enforce a canonical factorization, and two adapters minimizing their respective task 
losses may arrive at solutions expressed in entirely different latent coordinate 
systems. Under such representational misalignment, merging $\mathbf{A}^{(t)}$ with 
$\mathbf{A}^{(t-1)}$, or $\mathbf{B}^{(t)}$ with $\mathbf{B}^{(t-1)}$, conflates 
directions that are geometrically incomparable, introducing interference that stems 
not from task conflict, but from coordinate mismatch. This motivates the need for a 
principled approach that resolves the latent ambiguity before any consolidation is 
attempted. In subsequent sections, we lay the foundations for a principled approach to merge LoRA factors that respects the underlying coordinate systems.

\subsection{Stiefel Constraint as a Precondition for Alignment}
\label{subsec:stiefel_a}

We begin by addressing the ambiguity in the input factor $\mathbf{A}$, since it defines the latent coordinate system in which task-specific adaptations are represented. In unconstrained settings, this coordinate system is not identifiable: for any invertible matrix $\mathbf{R} \in GL(r)$, the factorization satisfies
\[
\mathbf{B}\mathbf{A} = (\mathbf{B}\mathbf{R})(\mathbf{R}^{-1}\mathbf{A}),
\]
implying that the same update can be represented in arbitrarily different coordinate systems. As a result, adapters trained across tasks may adopt incompatible latent bases, making subsequent comparison and merging ill-posed.

To address this, we constrain $\mathbf{A}$ to lie on the row-Stiefel manifold,
\begin{equation}
    \mathbf{A}\mathbf{A}^{\top} = \mathbf{I}_r.
\end{equation}

This constraint reduces the symmetry of the factorization, as formalized below.

\begin{lem}\label{lemma:keylemma}[Key Lemma: Reduction of invariance]
   Let $\mathbf{A}_t, \mathbf{A}_{t+1} \in \mathbb{R}^{r \times n}$ be input factors learned for two consecutive tasks where $t\geq 1$, both satisfying the Stiefel constraint $\mathbf{A}_t \mathbf{A}_t^\top = \mathbf{A}_{t+1} \mathbf{A}_{t+1}^\top = \mathbf{I}_r$. Suppose there exists an invertible matrix $\mathbf{R} \in GL(r)$ such that
\[
\mathbf{A}_{t+1} = \mathbf{R}^{-1} \mathbf{A}_t.
\]
Then $\mathbf{R}$ must satisfy $\mathbf{R}\mathbf{R}^\top = \mathbf{I}$, i.e., $\mathbf{R} \in \mathcal{O}(r)$. 
\end{lem}
\paragraph{Proof.}
Using $\mathbf{A}_{t+1} \mathbf{A}_{t+1}^\top = \mathbf{I}$, we obtain
\(
\mathbf{R}^{-1} \mathbf{A}_t \mathbf{A}_t^\top (\mathbf{R}^{-1})^\top = \mathbf{I}.
\)
Since $\mathbf{A}_{t} \mathbf{A}_{t}^\top = \mathbf{I}$, this reduces to
\(
\mathbf{R}^{-1} (\mathbf{R}^{-1})^\top = \mathbf{I},
\)
which implies $\mathbf{R}\mathbf{R}^\top = \mathbf{I}$.
\hfill $\square$




Lemma~\ref{lemma:keylemma} shows that the Stiefel constraint reduces the ambiguity of the LoRA factorization from arbitrary invertible transformations to orthogonal rotations. This is critical for continual merging: without it, alignment would require solving for a general linear transformation that may introduce scaling and shearing distortions. Restricting the ambiguity to $\mathcal{O}(r)$ instead yields a well-posed alignment problem that can be solved efficiently via Orthogonal Procrustes analysis (Section~\ref{subsec:b_alignment_motivation}).

We optimize $\mathbf{A}$ on the Stiefel manifold using standard Riemannian optimization methods (Appendix~\ref{app:stiefel_optimization}). Importantly, the role of the constraint is not to align adapters directly, but to impose a consistent geometric structure that makes subsequent alignment meaningful.

\subsection{Rotational Ambiguity and Alignment of \texorpdfstring{$\mathbf{B}$}{B}}
\label{subsec:b_alignment_motivation}

Constraining $\mathbf{A}$ to the Stiefel manifold removes general linear invariance, but the LoRA factorization remains ambiguous up to orthogonal transformations. For any $\mathbf{R}\in\mathcal{O}(r)$,
\(
\mathbf{B}\mathbf{A}
=
(\mathbf{B}\mathbf{R})
(\mathbf{R}^\top \mathbf{A}),
\)
so adapters may still differ by a rotation of their latent rank coordinates. While this ambiguity is irrelevant in single-task fine-tuning, it becomes problematic in continual learning, where factor matrices are merged across tasks. In particular, interpolation of output factors,
\(
\mathbf{B}^{(t)}_{\mathrm{merge}}
=
(1-\alpha)\mathbf{B}^{(t)}_{\mathrm{ft}}
+
\alpha \mathbf{B}^{(t-1)}_{\mathrm{merge}},
\)
is only meaningful when both factors are expressed in a common coordinate frame.

To resolve this mismatch, we align the previously merged factor
$\mathbf{B}_{\mathrm{old}}=\mathbf{B}^{(t-1)}_{\mathrm{merge}}$
to the coordinate system of the current task factor
$\mathbf{B}_{\mathrm{new}}=\mathbf{B}^{(t)}_{\mathrm{ft}}$
using Orthogonal Procrustes alignment \citep{procrustes}:
\begin{equation}\label{eq:procrustes optimization}
\mathbf{R}^*
=
\arg\min_{\mathbf{R}\in\mathcal{O}(r)}
\left\|
\mathbf{B}_{\mathrm{new}}
-
\mathbf{B}_{\mathrm{old}}\mathbf{R}^\top
\right\|_F^2.
\end{equation}

Let
\(
\mathbf{B}_{\mathrm{new}}^\top \mathbf{B}_{\mathrm{old}}
=
\mathbf{U}\mathbf{\Sigma}\mathbf{V}^\top
\)
be the singular value decomposition. The optimal rotation is given by \citep{procrustes}:
\(
\mathbf{R}^*=\mathbf{U}\mathbf{V}^\top,
\)
and the aligned factor is
\(
\tilde{\mathbf{B}}_{\mathrm{old}}
=
\mathbf{B}_{\mathrm{old}}(\mathbf{R}^*)^\top.
\)

This alignment step removes residual rotational mismatch while preserving the geometry of the learned subspace, enabling meaningful comparison and consolidation of rank directions across tasks. We adopt Orthogonal Procrustes because it provides a minimal, closed-form, and hyperparameter-free correction suitable for sequential continual learning.

\paragraph{Remark.}
Several prior works \citep{qiao2025mergeforgetsinglelora,zhu2024asymmetry,sun2024improving} have observed an asymmetry between the LoRA factors, where the input factor $\mathbf{A}$ tends to remain relatively stable across tasks while the output factor $\mathbf{B}$ captures a larger portion of task-specific adaptation. Our framework builds on this observation. Rather than explicitly merging $\mathbf{A}$ across tasks, we use the Stiefel constraint to stabilize its geometric structure and reduce factorization ambiguity. The primary consolidation mechanism is therefore applied to $\mathbf{B}$, whose representations are first aligned and then selectively merged across tasks. To further verify this, we conduct an additional ablation detailed in Appendix \ref{app:update_rule_ablations}

\subsection{From Global Averaging to per-rank retention gating}
\label{subsec:rankwise_b_merge}

While Procrustes alignment resolves the global coordinate mismatch between adapters, it does not determine how information should be consolidated across tasks. A natural approach is to interpolate between the aligned factors using a single global coefficient. However, such a strategy is inherently limited: different rank directions often evolve differently during fine-tuning. Some directions remain stable and encode information that is reusable across tasks, while others are substantially modified or repurposed for the current task.

This heterogeneity cannot be captured by a single interpolation weight. Instead, consolidation requires a mechanism to distinguish between directions that should retain prior knowledge and those that should follow the current update. Importantly, this decision depends on two complementary factors: whether the old and new directions are geometrically aligned, and how much the current task has altered the magnitude of the corresponding component.

To capture these effects, we introduce a rank-wise, forgetting-aware merging strategy that evaluates each direction independently. By combining directional agreement with a measure of magnitude drift, the method selectively retains prior information only where it remains compatible with the current task, while avoiding interference in directions that have been significantly rewritten.

\section{Methodology}

We now present SOAR, a continual LoRA learning framework consisting of two components:
(i) Stiefel-constrained optimization of the input factor $\mathbf{A}$ to stabilize the latent coordinate system, and
(ii) Procrustes-aligned, rank-wise consolidation of the output factor $\mathbf{B}$.
Algorithm~\ref{alg:soar} summarizes the overall procedure.

\paragraph{STEP 1: Stiefel optimization.}

We constrain $\mathbf{A}\in\mathbb{R}^{r\times n}$ to lie on the Stiefel manifold
\(
\mathrm{St}(r,n)
=
\{
\mathbf{A}\in\mathbb{R}^{r\times n}
\mid
\mathbf{A}\mathbf{A}^\top=\mathbf{I}_r
\}.
\)
Rather than enforcing orthogonality through penalty terms, we optimize directly on the manifold using Riemannian Adam (Appendix~\ref{app:stiefel_optimization}).

Although one could project $\mathbf{A}$ onto the manifold after fine-tuning, post-hoc projection breaks the coupling between $\mathbf{A}^{(t)}_{\mathrm{ft}}$ and $\mathbf{B}^{(t)}_{\mathrm{ft}}$, which are jointly optimized during LoRA adaptation (Appendix~\ref{app:post_hoc}).

\paragraph{STEP 2: Geometrically aligned merging.}
\label{para:b_merge}

Following the asymmetric roles of the LoRA factors, we directly update
$\mathbf{A}^{(t)}_{\mathrm{merge}} \leftarrow \mathbf{A}^{(t)}_{\mathrm{ft}}$
and perform continual consolidation only on $\mathbf{B}$.
Given the newly fine-tuned factor
$\mathbf{B}^{(t)}_{\mathrm{ft}}$
and the previously merged factor
$\mathbf{B}^{(t-1)}_{\mathrm{merge}}$,
our goal is to construct a consolidated output factor that preserves prior knowledge while adapting to the current task.
To this end, we first align the factors via Orthogonal Procrustes analysis and then apply rank-wise retention gating.

\paragraph{Procrustes alignment and per-rank retention gating}
\label{para:rankwise_b_merge}

Let $\mathbf{B}_{\mathrm{new}}=\mathbf{B}^{(t)}_{\mathrm{ft}}$ and $\mathbf{B}_{\mathrm{old}}=\mathbf{B}^{(t-1)}_{\mathrm{merge}}$. As discussed in Sec. \ref{subsec:b_alignment_motivation}, the Stiefel constraint reduces the LoRA ambiguity to rotations within the rank space. Before comparing or merging columns, we therefore align $\mathbf{B}_{\mathrm{old}}$ with the coordinate frame of $\mathbf{B}_{\mathrm{new}}$ using an orthogonal Procrustes transformation.

\paragraph{Procrustes alignment.}
As mentioned in Sec. \ref{subsec:b_alignment_motivation}, we solve the orthogonal Procrustes problem in Eq. (\ref{eq:procrustes optimization}), which is efficiently done by computing the thin SVD $\mathbf{B}_{\mathrm{new}}^{\top}\mathbf{B}_{\mathrm{old}}=\mathbf{U}\boldsymbol{\Sigma}\mathbf{V}^{\top}$.
This gives the closed-form solution $\mathbf{R}^{*}=\mathbf{U}\mathbf{V}^{\top}$. The aligned historical factor is then $\widetilde{\mathbf{B}}_{\mathrm{old}}=\mathbf{B}_{\mathrm{old}}(\mathbf{R}^{*})^{\top}$, expressed in the coordinate frame of $\mathbf{B}_{\mathrm{new}}$.

\paragraph{Per-rank retention gating.}
After aligning the old output factor with the new one, the remaining challenge is how to consolidate them. A single global interpolation weight is too coarse, since different rank directions evolve differently across tasks. Some directions remain stable and transferable, while others are substantially rewritten during adaptation. To capture this heterogeneity, we assign a separate retention score to each rank direction.

For each column $i\in\{1,\ldots,r\}$, let $\mathbf{b}_{\mathrm{new},i}$ denote the $i$-th column of $\mathbf{B}_{\mathrm{new}}$ and $\widetilde{\mathbf{b}}_{\mathrm{old},i}$ denote the $i$-th column of $\widetilde{\mathbf{B}}_{\mathrm{old}}$. We first measure their directional agreement:
\begin{equation}
    c_i
    =
    \frac{
    \mathbf{b}_{\mathrm{new},i}^{\top}
    \widetilde{\mathbf{b}}_{\mathrm{old},i}
    }{
    \|\mathbf{b}_{\mathrm{new},i}\|_2
    \|\widetilde{\mathbf{b}}_{\mathrm{old},i}\|_2
    + \epsilon
    }
    \in (-1,1),
    \label{eq:cosine_score}
\end{equation}
which captures whether the two adapters utilize a compatible output direction for rank component $i$.

Directional agreement alone is insufficient, since two columns may point in similar directions while differing substantially in magnitude. We therefore also compute the normalized residual
\begin{equation}
    \delta_i
    =
    \frac{
    \|\mathbf{b}_{\mathrm{new},i}
    -
    \widetilde{\mathbf{b}}_{\mathrm{old},i}\|_2^2
    }{
    \|\widetilde{\mathbf{b}}_{\mathrm{old},i}\|_2^2+\epsilon
    },
    \label{eq:forgetting_score}
\end{equation}
which quantifies how far the current task has moved the aligned historical direction relative to its original scale. These two signals are combined into a rank-wise retention gate:
\begin{equation}
    \gamma_i
    =
    \max(c_i,0)
    \exp\left(-({\delta_i}/{\tau})\right),
    \qquad
    \gamma_i\in[0,1],
    \label{eq:rankwise_gate}
\end{equation}
where $\tau>0$ controls sensitivity to magnitude drift. The gate assigns high retention only to directions that remain both positively aligned and minimally altered.

Letting
$\boldsymbol{\Gamma}=\mathrm{diag}(\gamma_1,\ldots,\gamma_r)$, we update the output factor as
\begin{equation}
    \mathbf{B}^{(t)}_{\mathrm{merge}}
    =
    \mathbf{B}_{\mathrm{new}}
    +
    \left(
    \widetilde{\mathbf{B}}_{\mathrm{old}}
    -
    \mathbf{B}_{\mathrm{new}}
    \right)
    \boldsymbol{\Gamma},
    \label{eq:additive_b_merge}
\end{equation}
or equivalently, for each rank direction,
\(
    \mathbf{b}^{(t)}_{\mathrm{merge},i}
    =
    (1-\gamma_i)\mathbf{b}_{\mathrm{new},i}
    +
    \gamma_i \widetilde{\mathbf{b}}_{\mathrm{old},i}.
    \)

This anchors consolidation at the current task update: when evidence for recovery is weak ($\gamma_i\approx0$), the new direction is preserved unchanged; when compatibility is strong ($\gamma_i\approx1$), prior information is selectively restored.

\paragraph{Catastrophic forgetting mitigation.}
The retention gate provides SOAR with an explicit mechanism for catastrophic forgetting control within the low-rank subspace. The cosine term $c_i$ measures directional compatibility between past and current adaptations, encouraging retention only when the corresponding rank directions remain aligned across tasks. The residual term $\delta_i$ measures the magnitude of task-induced change, suppressing retention for directions that have been substantially rewritten during adaptation. Together, these signals allow SOAR to selectively preserve reusable knowledge while permitting controlled forgetting in directions that have become task-specific, yielding a balance between stability and plasticity.

\section{Experiments}
\textbf{Models and Datasets. } We evaluate our approach across two model families and 3 models in total: Qwen3-8B, Llama-3.1-8B-Instruct, Llama-3.2-3B-Instruct. All experiments are conducted on one H100 GPU of 80GB. For evaluation similar to \citep{qiao2025mergeforgetsinglelora}, we consider 3 continual learning benchmarks: (1) \textbf{Standard CL benchmark: } DBpedia, AG News, Amazon Polarity, Yahoo Answers. \citep{zhang2016characterlevelconvolutionalnetworkstext}(2) \textbf{Large number of tasks: } It contains 4 standard CL tasks, along with GLUE tasks \citep{wang2019glue} (MNLI, QQP, SST-2, RTE) and 4 SuperGLUE tasks \citep{wang2019superglue} (WiC, ReCoRD, MultiRC, BoolQ) and IMDB movie reviews. Each of these tasks uses 2500 randomly sampled training samples and 500 test samples per class (3) \textbf{SuperNI Benchmark} \citep{wang2022supernaturalinstructionsgeneralizationdeclarativeinstructions}. It includes an extensive collection of NLP tasks with expert-written instructions, covering dialogue generation, summarization, sentiment analysis, question answering, and information extraction. For the SuperNI Benchmark, we use 1,000 training instances and 100 for validation/testing per task. To analyze the efficacy of our approach, as in previous work, we train models on eight different orders, which are task permutations. More details on task orders are detailed in Appendix (Table \ref{tab:orders}).

\textbf{Baselines.} We compare SOAR against different types of baselines: (1) Continual learning baselines: SLAO \citep{qiao2025mergeforgetsinglelora}, current benchmark for single LoRA-based continual learning method; SeqLoRA: Sequentially fine-tunes a single LoRA on multiple tasks without any modification; IncLoRA: Incrementally
adds a new LoRA per task while freezing previous LoRAs; O-LoRA \citep{wang2023orthogonalsubspacelearninglanguage}; GainLoRA \citep{liang2025gatedintegrationlowrankadaptation}; CorDA \citep{yang2025cordacontextorienteddecompositionadaptation}; (2) LoRA merging baselines: LoRA-LEGO \citep{zhao2024mergingloraslikeplaying}; KnOTS \citep{stoica2024modelmergingsvdtie}.

\textbf{Evaluation. } To evaluate the learning performance, we consider the following two metrics, i.e., accuracy (ACC), which measures the final accuracy averaged over all tasks, and backward transfer (BWT), which measures the average accuracy change of each task after learning new tasks:

\(
\mathrm{ACC} = \frac{1}{T} \sum_{i=1}^{T} A_{T,i}, 
\mathrm{BWT} = \frac{1}{T-1} \sum_{i=1}^{T-1} \left( A_{T,i} - A_{i,i} \right)
\), where $A_{i,j}$ represents the testing accuracy of task $j$ after learning task $i$.

\textbf{Experimental Setup. } All the experiments in the Table \ref{tab:main_tab} are performed on 1 H100, 80GB GPU with LoRA rank $\to$ 32, alpha $\to$ 64, learning rate $\to$ 5 $e^{-5}$. The $\tau$ to control the sensitivity of the gate in SOAR is set to 0.5 for the experiments. The targeted modules for LoRA training are q, k, v, and o. All the experiments are run 3 times on seeds $\{42, 43, 44\}$.

\begin{table*}[t]
\centering
\caption{(\%) Performance comparison on three CL benchmark datasets using  Qwen3-8B, across different task order, where $O_i$ represents $i^{th}$ task order.}
\small
\resizebox{\textwidth}{!}{%
\begin{tabular}{l|cccc|cccc|ccc}
\toprule
 & \multicolumn{4}{c|}{Standard CL Benchmark} 
 & \multicolumn{4}{c|}{Large Number of Tasks} 
 & \multicolumn{3}{c}{SuperNI Benchmark} \\
\midrule
Method 
& O1 $\uparrow$ & O2 $\uparrow$ & O3 $\uparrow$ & avg $\uparrow$ 
& O4 $\uparrow$ & O5 $\uparrow$ & O6 $\uparrow$ & avg $\uparrow$
& O7 $\uparrow$ & O8 $\uparrow$ & avg $\uparrow$ \\
\midrule
SeqLoRA    & 77.92& 78.05& 78.10& 78.02& 61.10& 61.33& 61.15& 61.19& 29.44& 31.43& 30.43\\
IncLoRA    & 82.20& 82.40& 82.92& 82.50& 65.20& 65.50& 65.40& 65.36& 33.44& 35.39& 34.42\\
O-LoRA     & 83.30& 83.20& 83.35& 83.28& 67.18& 67.08& 67.35& 67.20& 34.85& 34.66& 34.76\\
Gain-LoRA& 83.27& 83.36& 83.80& 83.48& 67.39& 67.39& 67.41& 67.39& 36.42& 37.38& 36.90\\
CorDA& 83.30& 83.30& 83.15& 83.25& 67.06& 67.12& 67.13& 67.10& 30.01& 30.66& 30.33\\
 SLAO& 85.25& 85.75& 86.98& 85.99& 68.19& 68.78& 68.51& 68.49& 47.72& 47.96&47.84\\
\midrule
KnOTS      & 76.60& 77.10& 76.70& 76.80& 59.23& 59.54& 59.46& 59.41& 26.99& 27.07& 27.03\\
LoRA-LEGO  & 82.63& 82.60& 82.70& 82.64& 67.72& 67.91& 66.64& 67.42& 40.95& 41.08& 41.02\\
\midrule
SOAR (ours)& \textbf{89.35}&\textbf{89.95}&\textbf{89.15}& 
           \textbf{89.48}& \textbf{71.23}&\textbf{70.93}&\textbf{70.72}& 
           \textbf{70.96}& \textbf{51.28}&\textbf{51.90}& \textbf{51.59}\\
\bottomrule
\end{tabular}}

\label{tab:main_tab}
\end{table*}

\subsection{Discussion}

\textbf{\textit{LoRA-based continual learning.}} As shown in Table~\ref{tab:main_tab}, SOAR consistently outperforms all baselines across three benchmarks on Qwen3-8B. SeqLoRA performs the worst because unconstrained sequential fine-tuning of a single shared adapter leads to severe catastrophic forgetting. IncLoRA improves by freezing previous adapters and introducing basic subspace isolation, but its memory cost grows linearly with tasks, and its separation strategy remains naive. O-LoRA provides modest gains through orthogonality constraints on new updates, yet applies them globally and does not exploit the asymmetric roles of LoRA factors. Gain-LoRA further improves via task-specific gating over shared components, but still treats $\mathbf{A}$ and $\mathbf{B}$ symmetrically, making it susceptible to coordinate drift during sequential merging. CorDA is competitive on the Standard CL and Large Number of Tasks benchmarks, but drops sharply on SuperNI, likely because nullspace selection from pretrained activations is less reliable under diverse instruction-following tasks. SLAO is the strongest LoRA-based baseline and benefits from single-adapter merging, but its factors are not aligned in a geometrically consistent coordinate system, limiting faithfulness of its consolidation step.


\textbf{\textit{LoRA merging baselines.}} From Table~\ref{tab:main_tab} we infer, KnOTS underperforms surprisingly relative to
continual learning methods across all three benchmarks, since it is designed for offline
model merging where all task adapters are available simultaneously and does not account for
the sequential, data-free nature of the continual setting. LoRA-LEGO performs more competitively
on the Standard CL and Large Number of Tasks benchmarks by treating LoRA modules as composable
blocks, but treats all rank directions equally and lacks any forgetting-aware consolidation,
which limits its performance on the SuperNI benchmark where task diversity is highest.
SOAR surpasses all baselines by significant margins across all settings, reaching $89.48$,
$70.96$, and $51.59$ average accuracy on the three benchmarks, respectively, validating that
jointly enforcing Stiefel-constrained input subspaces and Procrustes-aligned rank-wise merging
provides a principled and effective foundation for single-adapter continual LoRA learning. We report the mean and std for experiments in Table~\ref{tab:main_tab} in Appendix Figure~\ref{fig:var_main_fig}.

\begin{table*}[t]
\centering
\caption{(\%) Performance comparison of different initialization and merging on Qwen3-8B.}
\small
\resizebox{\textwidth}{!}{%
\begin{tabular}{l|cccc|cccc|ccc}
\toprule
 & \multicolumn{4}{c|}{Standard CL Benchmark} 
 & \multicolumn{4}{c|}{Large Number of Tasks} 
 & \multicolumn{3}{c}{SuperNI Benchmark} \\
\midrule
Method 
& O1 $\uparrow$& O2 $\uparrow$& O3 $\uparrow$& avg $\uparrow$
& O4 $\uparrow$& O5 $\uparrow$& O6 $\uparrow$& avg $\uparrow$
& O7 $\uparrow$& O8 $\uparrow$& avg $\uparrow$\\
\midrule
QR+lambda& 85.25& 85.75& 86.98& 85.99& 68.19& 68.78& 68.507& 68.49& 47.72& 47.96& 47.84\\
Stiefel+lambda& 86.10& 85.80& 86.75& 86.21& 69.05& 69.38& 68.96& 69.13& 48.35& 48.43& 48.40\\
 QR+SOAR Merging& 87.80& 87.50& 87.70& 87.67& 70.17& 70.01& 69.88& 70.02& 48.91& 49.68&49.29\\
\midrule
SOAR& \textbf{89.35}&\textbf{89.95}&\textbf{89.15}& 
           \textbf{89.48}& \textbf{71.23}&\textbf{70.93}&\textbf{70.72}& 
           \textbf{70.96}& \textbf{51.28}&\textbf{51.90}& \textbf{51.59}\\
\bottomrule
\end{tabular}}

\label{tab:init_merge}
\end{table*}


\paragraph{\textit{Ablation: Initialization and Merging Strategy.}}

Table~\ref{tab:init_merge} gives an ablation study over the two main components in SOAR: Stiefel initialization of $\mathbf{A}$, and Procrustes-aligned adaptive merging of $\mathbf{B}$. We evaluate four settings below: $\lambda$ based merging (same as SLAO \citep{qiao2025mergeforgetsinglelora}) is $\mathbf{B}_{\mathrm{merge}}$ = $\mathbf{B}_{\mathrm{old}}$ + $(1/\lambda)$ * ($\mathbf{B}_{\mathrm{new}}$ - $\mathbf{B}_{\mathrm{old}}$), where $\lambda$ is the square root of the task index. Here we show results of SOAR across three continual learning benchmarks on Qwen3-8B. First, replacing QR-based initialization with Stiefel optimization ({\it Stiefel+$\lambda$}) provides a slight improvement over {\it QR+$\lambda$} (e.g., $86.21$ vs.\ $85.99$ on the Standard CL Benchmark). This suggests that enforcing $\mathbf{A}^{\top}\in\mathrm{St}(n,r)$, equivalently $\mathbf{A}\mathbf{A}^{\top}=\mathbf{I}_r$, during training helps reduce the factorization ambiguity from $GL(r)$ to $\mathcal{O}(r)$, as shown in Lemma~\ref{lemma:keylemma}. However, without an alignment-aware merging rule for $\mathbf{B}$, the gain remains limited.
A larger benefit is observed for {\it QR+SOAR Merging}, showing that rotating $\mathbf{B}_{\mathrm{merge}}^{(t-1)}$ into the frame of $\mathbf{B}_{\mathrm{ft}}^{(t)}$ with $\mathbf{R}^{*} = \mathbf{U}\mathbf{V}^\top$, then gatedly preserving only those rank directions via $\gamma_i$ is indeed much more effective than uniform linear blending. 
Crucially, the quality of $\mathbf{R}^{*}$ depends on the geometric conditioning of the
input subspace: post-hoc QR projection onto Stiefel breaks the $\mathbf{A}$--$\mathbf{B}$
coupling established during training, causing alignment errors to compound across tasks (Appendix~\ref{app:post_hoc}).
By enforcing the Stiefel constraint \emph{during} training, SOAR ensures a well-conditioned
$\mathbf{R}^{*}$ and geometrically meaningful scores $c_i$ and $\delta_i$, explaining the consistent
margin over all ablated variants and confirming that Stiefel initialization is a
\emph{precondition} for well-posed Procrustes merging, not merely a complementary component.

\paragraph{\textit{Ablation: Effect of Procrustes and Per-rank Gate-based merging.}}

Figure~\ref{fig:b_ablation} reports the Backward Transfer (BWT) for three $\mathbf{B}$
merging strategies with $\mathbf{A}$ fixed on the Stiefel manifold, across task order, across all three settings, the trend is consistent and mirrors the theoretical predictions:
$\lambda$-merge exhibits the worst BWT (most negative), Procrustes+$\lambda$ improves over it, and SOAR achieves the least forgetting throughout.
The benefit of Procrustes alignment over plain $\lambda$-merge is visible even with a uniform interpolation coefficient: on (O7--O8), alignment alone recovers
$0.52$ BWT points ($-5.10 \to -4.58$), confirming that coordinate misalignment between
$\mathbf{B}_{\text{merge}}^{(t-1)}$ and $\mathbf{B}_{\text{ft}}^{(t)}$ introduces spurious interference that is independent of task conflict; Procrustes alignment removes this confound and allows the uniform merge to operate on geometrically comparable factors. However, the gap between Procrustes+$\lambda$ and SOAR is substantially larger and grows with task order ($0.35$ on O1--O3, $1.60$ on O4--O6, $1.67$ on O7--O8), revealing the fundamental limitation of uniform merging: as task sequences grow longer and more diverse, applying a single global coefficient to all rank directions increasingly over-retains directions that have been genuinely repurposed by the new task while under-protecting directions that are stable and compatible.

\begin{table}[t]
\centering
\small
\caption{Performance comparison of different model sizes and families.}
\label{tab:clean_table}
\resizebox{0.9\textwidth}{!}{%
\begin{tabular}{llccccccccccc}
\toprule
 & & \multicolumn{4}{c}{Standard CL Benchmark} 
   & \multicolumn{4}{c}{Large Number of Tasks} 
   & \multicolumn{3}{c}{SuperNI Benchmark} \\
\cmidrule(lr){3-6} \cmidrule(lr){7-10} \cmidrule(lr){11-13}

Model & Method 
& O1 $\uparrow$ & O2 $\uparrow$ & O3 $\uparrow$ & avg $\uparrow$
& O4 $\uparrow$ & O5 $\uparrow$ & O6 $\uparrow$ & avg $\uparrow$
& O7 $\uparrow$ & O8 $\uparrow$ & avg $\uparrow$\\
\midrule

\multirow{2}{*}{Llama-3.2-3B} 
& SLAO &  82.12 &  81.48 &  83.93&  82.51&  57.31&  62.13&  59.42&  59.92&  46.07&  45.52&  45.79\\
& SOAR &  \textbf{85.88}&  \textbf{86.45}&  \textbf{87.13}&  \textbf{86.48}&  \textbf{59.71}&  \textbf{63.31}&  \textbf{65.93}&  \textbf{62.98}&  \textbf{47.04}&  \textbf{48.87}&  \textbf{47.96}\\

\midrule
\multirow{2}{*}{Llama-3.1-8B} 
& SLAO &  87.08&  87.53&  87.35&  87.32&  65.25&  68.76&  65.80&  66.60&  47.50&  48.18&  47.84\\
& SOAR &  \textbf{89.35}&  \textbf{90.10}&  \textbf{89.70}&  \textbf{89.72}&  \textbf{68.42}&  \textbf{69.88}&  \textbf{68.49}&  \textbf{68.90}&  \textbf{49.98}&  \textbf{51.83}&  \textbf{50.91}\\

\midrule
\multirow{2}{*}{Qwen3-8B} 
& SLAO &  85.25&  85.75&  86.98&  85.99&  68.19&  68.78&  68.51&  68.49&  47.72&  47.96&  47.84\\
& SOAR &  \textbf{89.35}&  \textbf{89.95}&  \textbf{89.15}&  \textbf{89.48}&  \textbf{71.23}&  \textbf{70.93}&  \textbf{70.72}&  \textbf{70.96}&  \textbf{51.28}&  \textbf{51.90}&  \textbf{51.59}\\

\bottomrule
\end{tabular}}
\label{tab:model_fam}
\end{table}

\paragraph{\textit{Performance on different models of varying sizes.} }
Table~\ref{tab:model_fam} shows SOAR consistently outperforms SLAO across all model families
and sizes, with gains on SuperNI most pronounced for larger models ($+3.07$ and $+3.75$ for
Llama-3.1-8B and Qwen3-8B respectively), where task diversity makes faithful consolidation
critical. The consistency across architectures suggests the benefits of Stiefel-constrained
optimization and Procrustes-aligned merging scale with model expressivity rather than
being architecture-specific. More Detailed comparisons are in the Appendix Tables~\ref{tab:llama3.1_main_tab} and \ref{tab:llama3.2_main_tab}.

\paragraph{Additional ablations on gating:}
\label{app:add_ablations_gamma}

\begin{table}[ht]
\centering
\caption{Additional Ablation on Gating}
\begin{tabular}{lccccc}
\toprule
\textbf{Variant} & $\gamma_i$ & \textbf{O1} & \textbf{O4} & \textbf{O7} & \textbf{Avg} $\mathbf{(\Delta)}$\\
\midrule
Random gate	& $\sim \mathcal{U}[0,1]$ & 80.10	& 62.55	& 41.59	& 61.41 (-9.21)\\
Cosine-only	& $\max(c_i,0)$ & 86.05 & 65.33 & 44.41 & 65.26 (-5.36)\\
Residual-only & $\exp(-\delta_i/\tau)$ & 86.65 & 68.41 & 49.22 & 68.09 (-2.53)\\
\textbf{SOAR (full)} & $\max(c_i,0)\exp(-\delta_i/\tau)$ & \textbf{89.35} & \textbf{71.23} & \textbf{51.28} & \textbf{70.62}\\
\bottomrule
\end{tabular}

\label{tab:gamma_ablation}
\end{table}

All variants share the identical merge skeleton and Procrustes alignment, differing only in $\gamma_i$, so the comparison isolates the retention rule itself.

The ablation in Table~\ref{tab:gamma_ablation} confirms that the retention gate is a key component of SOAR. A random rank-wise gate performs substantially worse, demonstrating that continual learning performance depends on preserving the appropriate rank directions rather than simply retaining an equivalent amount of prior information. Furthermore, neither of the two signals used in the gate is sufficient on its own. Both the cosine-only and residual-only variants consistently underperform the full formulation across all benchmark suites, supporting the multiplicative design introduced in Eq. (\ref{eq:rankwise_gate}).

The cosine similarity captures directional agreement between old and new factors, while the residual term measures how much a rank direction has actually changed during adaptation. Relying solely on directional similarity may preserve directions whose magnitudes have already been significantly rewritten by the new task, whereas relying solely on residual magnitude ignores directional consistency. By combining both signals multiplicatively, the gate suppresses retention whenever either criterion indicates that a rank direction is no longer reliable, resulting in consistently stronger continual learning performance.

\paragraph{\textit{Additional Experiments.}} To further validate SOAR, we conduct additional experiments reported in the Appendix.
(1)~\textbf{Effect of rank} \ref{app:r_effect}: smaller ranks incur modest
degradation on the Large Number of Tasks and SuperNI benchmarks, while excessively large
ranks lead to overfitting, with $r{=}32$ providing the best trade-off.
(2)~\textbf{Effect of} \ref{app:tau_effect} $\tau$: smaller values of $\tau$
yield stronger performance on the more challenging benchmarks, as tighter drift penalization
better protects directions that have been substantially overwritten.
(3)~\textbf{Backward Transfer} \ref{app:bwt_comp}: SOAR achieves the least
negative BWT across all baselines, confirming that rank-wise gated merging provides
superior knowledge retention throughout the task sequence.
(4)~\textbf{Compute overhead} \ref{app:over_head}: SOAR incurs only $15.6\%$
additional wall-clock time over sequential fine-tuning while delivering substantially
larger accuracy gains, demonstrating a highly favorable efficiency performance trade-off.
(5)~\textbf{Ablation on Merging $\mathbf{A}$} \ref{app:update_rule_ablations}: Here we conduct experiments where we also merge $\mathbf{A}$ and we report the average change in $\mathbf{A}$ on varying tasks.
(6) ~\textbf{Gating Analysis} \ref{app:detail_gamma}: We also detail how the per-rank retention gating value varies across different attention.


\section{Conclusion}
We introduced SOAR, a geometrically based framework for continual LoRA learning that combines task updates into one adapter. By placing $\mathbf{A}$ on the Stiefel manifold and aligning $\mathbf{B}$ using Orthogonal Procrustes, SOAR minimizes coordinate confusion and allows for better rank-wise merging. In various continual learning tests, SOAR consistently enhances stability and flexibility while keeping adapter memory constant. One limitation is that SOAR retains information at the level of individual rank directions, which may not fully reflect knowledge spread across larger subspaces. Additionally, since SOAR does not use replay or saved data, information that gets completely overwritten during adaptation cannot be retrieved. Future efforts can build on this framework to achieve subspace-level consolidation and simpler memory solutions.

\medskip

\bibliography{final_version}

@inproceedings{qiao2025mergeforgetsinglelora,
 author = {Qiao, Fuli and Mahdavi, Mehrdad},
 booktitle = {International Conference on Learning Representations},
 pages = {8133--8159},
 title = {Merge before Forget: A Single LoRA Continual Learning via Continual Merging},
 url = {https://proceedings.iclr.cc/paper_files/paper/2026/file/0dedca10992d34035c9a9776254cfa8c-Paper-Conference.pdf},
 volume = {2026},
 year = {2026}
}

@inproceedings{wang2019superglue,
 author = {Wang, Alex and Pruksachatkun, Yada and Nangia, Nikita and Singh, Amanpreet and Michael, Julian and Hill, Felix and Levy, Omer and Bowman, Samuel},
 booktitle = {Advances in Neural Information Processing Systems},
 pages = {},
 publisher = {Curran Associates, Inc.},
 title = {Super{GLUE}: A Stickier Benchmark for General-Purpose Language Understanding Systems},
 url = {https://proceedings.neurips.cc/paper_files/paper/2019/file/4496bf24afe7fab6f046bf4923da8de6-Paper.pdf},
 volume = {32},
 year = {2019}
}

@inproceedings{wang2019glue,
  author       = {Alex Wang and
                  Amanpreet Singh and
                  Julian Michael and
                  Felix Hill and
                  Omer Levy and
                  Samuel R. Bowman},
  title        = {{GLUE:} {A} Multi-Task Benchmark and Analysis Platform for Natural
                  Language Understanding},
  booktitle    = {7th International Conference on Learning Representations, {ICLR} 2019,
                  New Orleans, LA, USA, May 6-9, 2019},
  year         = {2019},
  url          = {https://openreview.net/forum?id=rJ4km2R5t7}
}

@inproceedings{wang2022supernaturalinstructionsgeneralizationdeclarativeinstructions,
    title = "Super-{N}atural{I}nstructions: Generalization via Declarative Instructions on 1600+ {NLP} Tasks",
    author = "Wang, Yizhong  and
      Mishra, Swaroop  and
      Alipoormolabashi, Pegah  and
      Kordi, Yeganeh  and
      Mirzaei, Amirreza  and
      Naik, Atharva  and
      Ashok, Arjun  and
      Dhanasekaran, Arut Selvan  and
      Arunkumar, Anjana  and
      Stap, David  and
      Pathak, Eshaan  and
      Karamanolakis, Giannis  and
      Lai, Haizhi  and
      Purohit, Ishan  and
      Mondal, Ishani  and
      Anderson, Jacob  and
      Kuznia, Kirby  and
      Doshi, Krima  and
      Pal, Kuntal Kumar  and
      Patel, Maitreya  and
      Moradshahi, Mehrad  and
      Parmar, Mihir  and
      Purohit, Mirali  and
      Varshney, Neeraj  and
      Kaza, Phani Rohitha  and
      Verma, Pulkit  and
      Puri, Ravsehaj Singh  and
      Karia, Rushang  and
      Doshi, Savan  and
      Sampat, Shailaja Keyur  and
      Mishra, Siddhartha  and
      Reddy A, Sujan  and
      Patro, Sumanta  and
      Dixit, Tanay  and
      Shen, Xudong",
    booktitle = "Proceedings of the 2022 Conference on Empirical Methods in Natural Language Processing",
    month = dec,
    year = "2022",
    address = "Abu Dhabi, United Arab Emirates",
    publisher = "Association for Computational Linguistics",
    url = "https://aclanthology.org/2022.emnlp-main.340/",
    doi = "10.18653/v1/2022.emnlp-main.340",
    pages = "5085--5109"
}

@inproceedings{wang2023orthogonalsubspacelearninglanguage,
    title = "Orthogonal Subspace Learning for Language Model Continual Learning",
    author = "Wang, Xiao  and
      Chen, Tianze  and
      Ge, Qiming  and
      Xia, Han  and
      Bao, Rong  and
      Zheng, Rui  and
      Zhang, Qi  and
      Gui, Tao  and
      Huang, Xuanjing",
    booktitle = "Findings of the Association for Computational Linguistics: EMNLP 2023",
    month = dec,
    year = "2023",
    address = "Singapore",
    publisher = "Association for Computational Linguistics",
    url = "https://aclanthology.org/2023.findings-emnlp.715/",
    doi = "10.18653/v1/2023.findings-emnlp.715",
    pages = "10658--10671"
}

@inproceedings{liang2025gatedintegrationlowrankadaptation,
 author = {Liang, Yan-Shuo and Chen, Jia-Rui and Li, Wu-Jun},
 booktitle = {Advances in Neural Information Processing Systems},
 doi = {10.52202/085713-2310},
 pages = {68659--68689},
 publisher = {Curran Associates, Inc.},
 title = {Gated Integration of Low-Rank Adaptation for Continual Learning  of Large Language Models},
 url = {https://proceedings.neurips.cc/paper_files/paper/2025/file/63692d8d567db671c700df5df912204a-Paper-Conference.pdf},
 volume = {38, Main Conference},
 year = {2025}
}

@inproceedings{yang2025cordacontextorienteddecompositionadaptation,
 author = {Yang, Yibo and Li, Xiaojie and Zhou, Zhongzhu and Song, Shuaiwen Leon and Wu, Jianlong and Nie, Liqiang and Ghanem, Bernard},
 booktitle = {Advances in Neural Information Processing Systems},
 doi = {10.52202/079017-2292},
 pages = {71768--71791},
 publisher = {Curran Associates, Inc.},
 title = {CorDA: Context-Oriented Decomposition Adaptation of Large Language Models for Task-Aware Parameter-Efficient Fine-tuning},
 url = {https://proceedings.neurips.cc/paper_files/paper/2024/file/83f95bb0ac5046338ea2afe3390e9f4b-Paper-Conference.pdf},
 volume = {37},
 year = {2024}
}

@inproceedings{zhao2024saptsharedattentionframework,
    title = "{SAPT}: A Shared Attention Framework for Parameter-Efficient Continual Learning of Large Language Models",
    author = "Zhao, Weixiang  and
      Wang, Shilong  and
      Hu, Yulin  and
      Zhao, Yanyan  and
      Qin, Bing  and
      Zhang, Xuanyu  and
      Yang, Qing  and
      Xu, Dongliang  and
      Che, Wanxiang",
    booktitle = "Proceedings of the 62nd Annual Meeting of the Association for Computational Linguistics (Volume 1: Long Papers)",
    month = aug,
    year = "2024",
    address = "Bangkok, Thailand",
    publisher = "Association for Computational Linguistics",
    url = "https://aclanthology.org/2024.acl-long.625/",
    doi = "10.18653/v1/2024.acl-long.625",
    pages = "11641--11661"
}

@inproceedings{zhao2024mergingloraslikeplaying,
 author = {Zhao, Ziyu and shen, tao and Zhu, Didi and Li, Zexi and Su, Jing and Wang, Xuwu and Wu, Fei},
 booktitle = {International Conference on Learning Representations},
 pages = {72896--72913},
 title = {Merging LoRAs like Playing LEGO: Pushing the Modularity of LoRA to Extremes Through Rank-Wise Clustering},
 url = {https://proceedings.iclr.cc/paper_files/paper/2025/file/b54e0146a82945f01e69c2e3309ba925-Paper-Conference.pdf},
 volume = {2025},
 year = {2025}
}

@inproceedings{stoica2024modelmergingsvdtie,
 author = {Stoica, George and Ramesh, Pratik and Ecsedi, Boglarka and Choshen, Leshem and Hoffman, Judy},
 booktitle = {International Conference on Learning Representations},
 pages = {4501--4519},
 title = {Model merging with SVD to tie the Knots},
 url = {https://proceedings.iclr.cc/paper_files/paper/2025/file/0d4f8a5109c5083b5307fcd0bddae7a7-Paper-Conference.pdf},
 volume = {2025},
 year = {2025}
}

@article{10.1145/3735633,
author = {Shi, Haizhou and Xu, Zihao and Wang, Hengyi and Qin, Weiyi and Wang, Wenyuan and Wang, Yibin and Wang, Zifeng and Ebrahimi, Sayna and Wang, Hao},
title = {Continual Learning of Large Language Models: A Comprehensive Survey},
year = {2025},
issue_date = {April 2026},
publisher = {Association for Computing Machinery},
address = {New York, NY, USA},
volume = {58},
number = {5},
issn = {0360-0300},
url = {https://doi.org/10.1145/3735633},
doi = {10.1145/3735633},
journal = {ACM Comput. Surv.},
month = nov,
articleno = {120},
numpages = {42}
}

@article{wang2024comprehensivesurveycontinuallearning,
  author={Wang, Liyuan and Zhang, Xingxing and Su, Hang and Zhu, Jun},
  journal={IEEE Transactions on Pattern Analysis and Machine Intelligence}, 
  title={A Comprehensive Survey of Continual Learning: Theory, Method and Application}, 
  year={2024},
  volume={46},
  number={8},
  pages={5362-5383},
  doi={10.1109/TPAMI.2024.3367329}}

@inproceedings{fa2026decomposingcomposingefficientvisionlanguage,
  title={Decomposing and Composing: Towards Efficient Vision-Language Continual Learning via Rank-1 Expert Pool in a Single LoRA},
  author={Fa, Zhan and Duan, Yue and Zhang, Jian and Qi, Lei and Yang, Wanqi and Shi, Yinghuan},
  booktitle={Proceedings of the AAAI Conference on Artificial Intelligence},
  volume={40},
  number={25},
  pages={20985--20993},
  year={2026},
  doi={10.1609/aaai.v40i25.39240},
  url={https://doi.org/10.1609/aaai.v40i25.39240}
}

@misc{han2026w2tloraweightsknow,
      title={W2T: LoRA Weights Already Know What They Can Do}, 
      author={Xiaolong Han and Ferrante Neri and Zijian Jiang and Fang Wu and Yanfang Ye and Lu Yin and Zehong Wang},
      year={2026},
      eprint={2603.15990},
      archivePrefix={arXiv},
      primaryClass={cs.LG},
      url={https://arxiv.org/abs/2603.15990}, 
}

@inproceedings{kumar2025latentspacefactorizationlora,
 author = {Kumar, Shashi and Kaloga, Yacouba and Mitros, John and Motlicek, Petr and Kodrasi, Ina},
 booktitle = {Advances in Neural Information Processing Systems},
 doi = {10.52202/085713-2945},
 pages = {88111--88138},
 publisher = {Curran Associates, Inc.},
 title = {Latent Space Factorization in LoRA},
 url = {https://proceedings.neurips.cc/paper_files/paper/2025/file/7f8662c03b26acab4eb94c66ba1aa7fc-Paper-Conference.pdf},
 volume = {38, Main Conference},
 year = {2025}
}

@inproceedings{liang2024inflora,
  author={Liang, Yan-Shuo and Li, Wu-Jun},
  booktitle={2024 IEEE/CVF Conference on Computer Vision and Pattern Recognition (CVPR)}, 
  title={{InfLoRA}: Interference-Free Low-Rank Adaptation for Continual Learning}, 
  year={2024},
  volume={},
  number={},
  pages={23638-23647},
  doi={10.1109/CVPR52733.2024.02231}
}

@InProceedings{hao2024floralowrankadapterssecretly,
  title = 	 {Flora: Low-Rank Adapters Are Secretly Gradient Compressors},
  author =       {Hao, Yongchang and Cao, Yanshuai and Mou, Lili},
  booktitle = 	 {Proceedings of the 41st International Conference on Machine Learning},
  pages = 	 {17554--17571},
  year = 	 {2024},
  volume = 	 {235},
  series = 	 {Proceedings of Machine Learning Research},
  month = 	 {21--27 Jul},
  publisher =    {PMLR},
  url = 	 {https://proceedings.mlr.press/v235/hao24a.html}
}

@inproceedings{rolnick2019experiencereplaycontinuallearning,
 author = {Rolnick, David and Ahuja, Arun and Schwarz, Jonathan and Lillicrap, Timothy and Wayne, Gregory},
 booktitle = {Advances in Neural Information Processing Systems},
 pages = {},
 publisher = {Curran Associates, Inc.},
 title = {Experience Replay for Continual Learning},
 url = {https://proceedings.neurips.cc/paper_files/paper/2019/file/fa7cdfad1a5aaf8370ebeda47a1ff1c3-Paper.pdf},
 volume = {32},
 year = {2019}
}

@article{Li_2024,
title = {{AdaER}: An adaptive experience replay approach for continual lifelong learning},
journal = {Neurocomputing},
volume = {572},
pages = {127204},
year = {2024},
issn = {0925-2312},
doi = {https://doi.org/10.1016/j.neucom.2023.127204},
url = {https://www.sciencedirect.com/science/article/pii/S0925231223013279},
author = {Xingyu Li and Bo Tang and Haifeng Li}
}

@inproceedings{lopezpaz2022gradientepisodicmemorycontinual,
 author = {Lopez-Paz, David and Ranzato, Marc\textquotesingle Aurelio},
 booktitle = {Advances in Neural Information Processing Systems},
 pages = {},
 publisher = {Curran Associates, Inc.},
 title = {Gradient Episodic Memory for Continual Learning},
 url = {https://proceedings.neurips.cc/paper_files/paper/2017/file/f87522788a2be2d171666752f97ddebb-Paper.pdf},
 volume = {30},
 year = {2017}
}

@article{
doi:10.1073/pnas.1611835114,
author = {James Kirkpatrick  and Razvan Pascanu  and Neil Rabinowitz  and Joel Veness  and Guillaume Desjardins  and Andrei A. Rusu  and Kieran Milan  and John Quan  and Tiago Ramalho  and Agnieszka Grabska-Barwinska  and Demis Hassabis  and Claudia Clopath  and Dharshan Kumaran  and Raia Hadsell },
title = {Overcoming catastrophic forgetting in neural networks},
journal = {Proceedings of the National Academy of Sciences},
volume = {114},
number = {13},
pages = {3521-3526},
year = {2017},
doi = {10.1073/pnas.1611835114},
URL = {https://www.pnas.org/doi/abs/10.1073/pnas.1611835114},
eprint = {https://www.pnas.org/doi/pdf/10.1073/pnas.1611835114}
}

@article{li2017learningforgetting,
  author={Li, Zhizhong and Hoiem, Derek},
  journal={IEEE Transactions on Pattern Analysis and Machine Intelligence}, 
  title={Learning without Forgetting}, 
  year={2018},
  volume={40},
  number={12},
  pages={2935-2947},
  doi={10.1109/TPAMI.2017.2773081}}

@INPROCEEDINGS{mallya2018packnetaddingmultipletasks,
  author={Mallya, Arun and Lazebnik, Svetlana},
  booktitle={2018 IEEE/CVF Conference on Computer Vision and Pattern Recognition}, 
  title={{PackNet}: Adding Multiple Tasks to a Single Network by Iterative Pruning}, 
  year={2018},
  volume={},
  number={},
  pages={7765-7773},
  doi={10.1109/CVPR.2018.00810}}

@InProceedings{serra2018overcomingcatastrophicforgettinghard,
  title = 	 {Overcoming Catastrophic Forgetting with Hard Attention to the Task},
  author =       {Serra, Joan and Suris, Didac and Miron, Marius and Karatzoglou, Alexandros},
  booktitle = 	 {Proceedings of the 35th International Conference on Machine Learning},
  pages = 	 {4548--4557},
  year = 	 {2018},
  volume = 	 {80},
  series = 	 {Proceedings of Machine Learning Research},
  month = 	 {10--15 Jul},
  publisher =    {PMLR},
  url = 	 {https://proceedings.mlr.press/v80/serra18a.html}
}

@inproceedings{ilharco2023editingmodelstaskarithmetic,
title={Editing models with task arithmetic},
author={Gabriel Ilharco and Marco T{\'{u}}lio Ribeiro and Mitchell Wortsman and Ludwig Schmidt and Hannaneh Hajishirzi and Ali Farhadi},
booktitle={The Eleventh International Conference on Learning Representations, {ICLR} 2023},
year={2023},
url={https://openreview.net/forum?id=6t0Kwf8-jrj}
}

@inproceedings{yadav2023tiesmergingresolvinginterferencemerging,
 author = {Yadav, Prateek and Tam, Derek and Choshen, Leshem and Raffel, Colin and Bansal, Mohit},
 booktitle = {Advances in Neural Information Processing Systems},
 doi = {10.52202/075280-0310},
 pages = {7093--7115},
 publisher = {Curran Associates, Inc.},
 title = {{TIES-Merging}: Resolving Interference When Merging Models},
 url = {https://proceedings.neurips.cc/paper_files/paper/2023/file/1644c9af28ab7916874f6fd6228a9bcf-Paper-Conference.pdf},
 volume = {36},
 year = {2023}
}

@misc{tang2025mergingmodelsflyretraining,
      title={Merging Models on the Fly Without Retraining: A Sequential Approach to Scalable Continual Model Merging}, 
      author={Anke Tang and Enneng Yang and Li Shen and Yong Luo and Han Hu and Bo Du and Dacheng Tao},
      year={2025},
      eprint={2501.09522},
      archivePrefix={arXiv},
      primaryClass={cs.LG},
      url={https://arxiv.org/abs/2501.09522}, 
}

@inproceedings{zhu2024asymmetry,
  title = 	 {Asymmetry in Low-Rank Adapters of Foundation Models},
  author =       {Zhu, Jiacheng and Greenewald, Kristjan and Nadjahi, Kimia and S\'{a}ez De Oc\'{a}riz Borde, Haitz and Gabrielsson, Rickard Br\"{u}el and Choshen, Leshem and Ghassemi, Marzyeh and Yurochkin, Mikhail and Solomon, Justin},
  booktitle = 	 {Proceedings of the 41st International Conference on Machine Learning},
  pages = 	 {62369--62385},
  year = 	 {2024},
  volume = 	 {235},
  series = 	 {Proceedings of Machine Learning Research},
  month = 	 {21--27 Jul},
  publisher =    {PMLR},
  url = 	 {https://proceedings.mlr.press/v235/zhu24c.html}
}

@inproceedings{sun2024improving,
 author = {Sun, Youbang and Li, Zitao and Li, Yaliang and Ding, Bolin},
 booktitle = {International Conference on Learning Representations},
 editor = {B. Kim and Y. Yue and S. Chaudhuri and K. Fragkiadaki and M. Khan and Y. Sun},
 pages = {17978--17994},
 title = {Improving {LoRA} in Privacy-preserving Federated Learning},
 url = {https://proceedings.iclr.cc/paper_files/paper/2024/file/4e243e95c913b367775d71d7182b99d9-Paper-Conference.pdf},
 volume = {2024},
 year = {2024}
}

@article{procrustes,
 title={A Generalized Solution of the Orthogonal Procrustes Problem},
 volume={31},
 DOI={10.1007/BF02289451},
 number={1},
 journal={Psychometrika},
 author={Sch{\"o}nemann, Peter H.},
 year={1966},
 pages={1–10},
 url={https://api.semanticscholar.org/CorpusID:121676935}
}

@inproceedings{zhang2016characterlevelconvolutionalnetworkstext,
 author = {Zhang, Xiang and Zhao, Junbo and LeCun, Yann},
 booktitle = {Advances in Neural Information Processing Systems},
 pages = {},
 publisher = {Curran Associates, Inc.},
 title = {Character-level Convolutional Networks for Text Classification},
 url = {https://proceedings.neurips.cc/paper_files/paper/2015/file/250cf8b51c773f3f8dc8b4be867a9a02-Paper.pdf},
 volume = {28},
 year = {2015}
}

@misc{yang2025qwen3technicalreport,
      title={Qwen3 Technical Report}, 
      author={An Yang and Anfeng Li and Baosong Yang and Beichen Zhang and Binyuan Hui and Bo Zheng and Bowen Yu and Chang Gao and Chengen Huang and Chenxu Lv and Chujie Zheng and Dayiheng Liu and Fan Zhou and Fei Huang and Feng Hu and Hao Ge and Haoran Wei and Huan Lin and Jialong Tang and Jian Yang and Jianhong Tu and Jianwei Zhang and Jianxin Yang and Jiaxi Yang and Jing Zhou and Jingren Zhou and Junyang Lin and Kai Dang and Keqin Bao and Kexin Yang and Le Yu and Lianghao Deng and Mei Li and Mingfeng Xue and Mingze Li and Pei Zhang and Peng Wang and Qin Zhu and Rui Men and Ruize Gao and Shixuan Liu and Shuang Luo and Tianhao Li and Tianyi Tang and Wenbiao Yin and Xingzhang Ren and Xinyu Wang and Xinyu Zhang and Xuancheng Ren and Yang Fan and Yang Su and Yichang Zhang and Yinger Zhang and Yu Wan and Yuqiong Liu and Zekun Wang and Zeyu Cui and Zhenru Zhang and Zhipeng Zhou and Zihan Qiu},
      year={2025},
      eprint={2505.09388},
      archivePrefix={arXiv},
      primaryClass={cs.CL},
      url={https://arxiv.org/abs/2505.09388}, 
}

@misc{grattafiori2024llama3herdmodels,
      title={{The Llama 3 Herd of Models}}, 
      author={Aaron Grattafiori and Abhimanyu Dubey and Abhinav Jauhri and others},
      year={2024},
      eprint={2407.21783},
      archivePrefix={arXiv},
      primaryClass={cs.AI},
      url={https://arxiv.org/abs/2407.21783}, 
}
\bibliographystyle{plain}
\appendix
\section{Related Works on General Continual Learning}
Continual learning has been a major interest in the deep learning community, aiming for lifelong learning and model evolution. Previous approaches have explored various methods (see \citep{wang2024comprehensivesurveycontinuallearning} and references therein) to tackle catastrophic forgetting and improve knowledge sharing: To tackle two major problems in continual learning tasks: (a) catastrophic forgetting, where model performance degrades on previous tasks when learning new tasks, and (b) knowledge sharing, where knowledge acquired from previous tasks is leveraged to improve learning on subsequent tasks. Some earlier methods explored are 1) Memory replay-based methods, such as: CLEAR \citep{rolnick2019experiencereplaycontinuallearning} using a replay buffer with mixed on/off-policy learning and behavioral cloning, which enables reinforcement learning agents to learn new tasks while significantly reducing catastrophic forgetting. AdaER \citep{Li_2024} improves replay through selective memory recall (conflict-based sampling) and entropy-based buffer management. GEM \citep{lopezpaz2022gradientepisodicmemorycontinual} stores task-specific gradients and projects new gradients to avoid interference with previous knowledge. 2) Regularization-based methods: EWC \citep{doi:10.1073/pnas.1611835114} penalizes changes to weights important for past tasks using a Fisher information–based quadratic constraint. LwF \citep{li2017learningforgetting} uses distillation loss to preserve old task predictions while learning new tasks. 3) Architecture-based methods: Packnet \citep{mallya2018packnetaddingmultipletasks} iteratively prunes and freezes weights to allocate capacity for new tasks. HAT \citep{serra2018overcomingcatastrophicforgettinghard} learns task-specific masks to gate which neurons are used.

\section{Optimization on the Stiefel Manifold}
\label{app:stiefel_optimization}

To enforce the orthogonality constraint $\mathbf{A}\mathbf{A}^\top = \mathbf{I}_r$, we perform optimization on the row-Stiefel manifold
\[
\mathrm{St}(r,n) = \left\{ \mathbf{A} \in \mathbb{R}^{r \times n} : \mathbf{A}\mathbf{A}^\top = \mathbf{I}_r \right\}.
\]

Given the Euclidean gradient $\mathbf{G}_{\mathbf{A}} = \nabla_{\mathbf{A}} \mathcal{L}$, the tangent space at $\mathbf{A}$ is
\[
T_{\mathbf{A}}\mathrm{St}(r,n) =
\left\{
\boldsymbol{\Xi} \in \mathbb{R}^{r \times n} :
\boldsymbol{\Xi}\mathbf{A}^\top + \mathbf{A}\boldsymbol{\Xi}^\top = \mathbf{0}
\right\}.
\]

The Riemannian gradient is obtained by projecting the Euclidean gradient onto the tangent space:
\[
\mathrm{grad}_{\mathbf{A}} \mathcal{L}
=
\mathbf{G}_{\mathbf{A}} - \mathrm{sym}(\mathbf{G}_{\mathbf{A}}\mathbf{A}^\top)\mathbf{A},
\quad
\mathrm{sym}(\mathbf{M}) = \tfrac{1}{2}(\mathbf{M} + \mathbf{M}^\top).
\]

After taking a step along a tangent direction $\boldsymbol{\Xi}$, we retract back to the manifold. We use the polar retraction:
\[
\mathrm{Retr}_{\mathbf{A}}(\boldsymbol{\Xi})
=
\left( \mathbf{I}_r + \boldsymbol{\Xi}\boldsymbol{\Xi}^\top \right)^{-1/2}
(\mathbf{A} + \boldsymbol{\Xi}),
\]
which ensures that the updated $\mathbf{A}$ remains orthonormal.

In practice, this procedure maintains $\mathbf{A}\mathbf{A}^\top = \mathbf{I}_r$ throughout training, rather than encouraging orthogonality via soft regularization. This is critical for our method, as it guarantees that the latent coordinate system remains well-conditioned and comparable across tasks, enabling the alignment procedure described in the main text.

\section{Impact of constraining $\mathbf{A}$ on Stiefel Manifold}
\paragraph{Effect of post-hoc projection on coupled factors.}
\label{app:post_hoc}
In LoRA, the factors $\mathbf{A}$ and $\mathbf{B}$ are jointly optimized, and the learned pair $(\mathbf{B}, \mathbf{A})$ represents a particular element within an equivalence class under transformations of the form $(\mathbf{B}, \mathbf{A}) \mapsto (\mathbf{B}\mathbf{R}^\top, \mathbf{R}^{-1}\mathbf{A})$ for $\mathbf{R} \in GL(r)$. As a result, the optimizer implicitly selects a coordinate system in which $\mathbf{A}$ and $\mathbf{B}$ are coupled.

If $\mathbf{A}$ is subsequently projected onto the Stiefel manifold to enforce $\mathbf{A}\mathbf{A}^\top = \mathbf{I}_r$ (by, for instance, QR-decomposition), this operation modifies the coordinate system of $\mathbf{A}$. However, unless a corresponding transformation is applied to $\mathbf{B}$, the original coupling between the two factors is broken, and the resulting pair no longer represents the same low-rank update.

Even if one attempts to compensate by transforming $\mathbf{B}$ after projecting $\mathbf{A}$, the required transformation generally lies in $GL(r)$, reflecting the unconstrained symmetry of the LoRA factorization during training. However, the subsequent alignment step is restricted to orthogonal transformations, as Procrustes solves an optimization over $\mathcal{O}(r)$. As a result, it can only recover the closest orthogonal approximation to the true transformation, but cannot capture general scaling or shearing components. This mismatch leads to residual misalignment between factors, as the coordinate discrepancy introduced during training is only partially corrected. In a continual setting, such approximation errors can accumulate across tasks, degrading the consistency of rank directions and affecting subsequent merging.

By enforcing the Stiefel constraint throughout training, we ensure that $\mathbf{A}$ and $\mathbf{B}$ are learned within a restricted equivalence class (up to orthogonal transformations), preserving their coupling and maintaining a consistent coordinate system across tasks.

\begin{table}[t]
\centering
\caption{Effect of different ranks on SOAR using Qwen3-8B.}

\small
\resizebox{\textwidth}{!}{%
\begin{tabular}{l|cccc|cccc|ccc}
\toprule
 & \multicolumn{4}{c|}{Standard CL Benchmark} 
 & \multicolumn{4}{c|}{Large Number of Tasks} 
 & \multicolumn{3}{c}{SuperNI Benchmark} \\
\midrule
Method 
& O1 $\uparrow$ & O2 $\uparrow$ & O3 $\uparrow$ & avg $\uparrow$
& O4 $\uparrow$ & O5 $\uparrow$ & O6 $\uparrow$ & avg $\uparrow$
& O7 $\uparrow$ & O8 $\uparrow$ & avg $\uparrow$\\
\midrule
$r \to 8$& 88.82& 88.88& 88.38& 88.69& 69.34& 69.07& 69.70& 69.37& 49.50& 50.09& 49.79\\
$r \to 16$& 88.78& 88.87& 88.60& 88.72& 70.20& 70.13& 70.19& 70.17& 49.58 & 50.83& 50.21\\
 $r \to 32$& \textbf{89.35}& \textbf{89.95}& \textbf{89.15}& \textbf{89.48}& 71.23& \textbf{70.93}& 70.72& \textbf{70.96}& 51.28& \textbf{51.90}&\textbf{51.59}\\
$r \to 64$& 88.60&88.78&89.13& 
           88.83& \textbf{71.40}&70.44&\textbf{70.80}& 
           70.88& \textbf{51.60}&51.42 & 51.51\\
\bottomrule
\end{tabular}}
\label{tab:alpha_effect}
\end{table}

\begin{table}[h]
\centering
\caption{BWT scores across task orders on different models.}
\resizebox{\textwidth}{!}{
\begin{tabular}{lrrrrrrrrr}
\toprule
\textbf{Method} & \textbf{A (SCL)} & \textbf{B (SCL)} & \textbf{C (SCL)} & \textbf{A (LNT)} & \textbf{B (LNT)} & \textbf{C (LNT)} & \textbf{A (SNI)} & \textbf{B (SNI)} & \textbf{C (SNI)}\\
\midrule
SeqLoRA & -12.40 & -12.85 & -13.62 & -14.41 & -14.78 & -15.81 & -17.57 & -18.04 & -19.42\\
IncLoRA & -8.02 & -8.74 & -9.38 & -8.41 & -9.12 & -10.21 & -14.98 & -15.46 & -16.84\\
O-LoRA & -7.10 & -7.58 & -8.16 & -5.66 & -6.18 & -6.91 & -13.78 & -14.02 & -15.18\\
Gain-LoRA & -7.35 & -7.96 & -8.42 & -4.54 & -5.08 & -5.84 & -11.39 & -11.87 & -12.73\\
CoRDA & -7.02 & -7.46 & -8.37 & -7.61 & -8.04 & -8.92 & -15.56 & -16.03 & -17.44\\
SLAO & -4.12 & -4.38 & -5.18 & -3.53 & -3.91 & -4.67 & -6.04 & -6.28 & -6.21\\
\midrule
KnoTS & -12.65 & -13.08 & -13.94 & -11.55 & -11.97 & -12.81 & -22.66 & -23.21 & -26.85\\
LoRA-LEGO & -7.98 & -8.41 & -9.05 & -7.51 & -7.96 & -8.73 & -9.89 & -10.54 & -11.62\\
\midrule
SOAR (ours) & \textbf{-1.25} & \textbf{-1.41} & \textbf{-1.86} & \textbf{-1.68} & \textbf{-1.92} & \textbf{-2.31} & \textbf{-3.43} & \textbf{-3.71} & \textbf{-4.26}\\
\bottomrule
\end{tabular}}

\label{tab:bwt_comparison}
\end{table}



\section{Further Experimental Analysis}
\subsection{Effect of LoRA ranks:}
\label{app:r_effect}
Table~\ref{tab:alpha_effect} looks at how LoRA rank affects SOAR's performance on Qwen3-8B. Performance improves steadily from $r=8$ to $r=32$ across all three benchmarks. The Standard CL and SuperNI benchmarks show the most noticeable trends ($88.69-89.48$ and $49.79-51.59$, respectively). This indicates that a larger rank gives more expressive low-rank directions for the Procrustes alignment and per-rank gating to work with. 
After $r=32$, the gains level off or even drop slightly. For example, $r=64$ achieves similar performance on the Large Number of Tasks benchmark ($70.88$ vs. $70.96$) but shows a small decrease on the Standard CL Benchmark ($88.83$ vs. $89.48$). This may indicate mild overfitting or redundancy in the extra rank directions. 
Overall, it seems that $r=32$ provides the best balance between subspace expressiveness and the geometric consistency required by the Stiefel constraint as the rank increases. Maintaining a well-conditioned orthonormal basis becomes more difficult, and the additional rank directions offer less coherent structure to the alignment process.

\begin{figure}[h]
    \centering
    \includegraphics[width=0.8\linewidth]{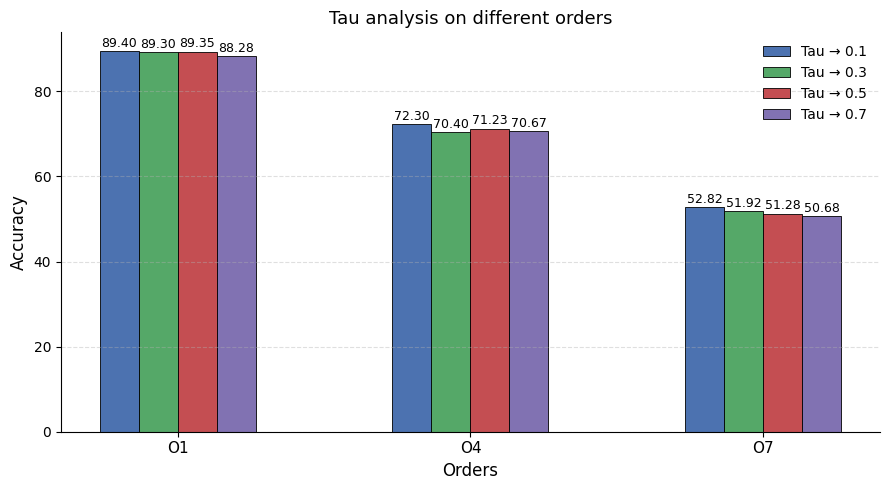}
    \caption{(\%) Performance analysis across different $\tau$ values for three task orders.}
    \label{fig:tau_ablation}
\end{figure}

\subsection{Ablation on the impact of $\tau$: }
\label{app:tau_effect}
Figure~\ref{fig:tau_ablation} examines the sensitivity of SOAR to the temperature parameter
$\tau$, which controls the sensitivity of the retention gate $\gamma_i = \max(c_i,0)
\cdot \exp(-\delta_i/\tau)$ to magnitude drift $\delta_i$.
Across all three benchmarks, performance is remarkably stable for $\tau \in \{0.1, 0.3, 0.5\}$,
with differences of at most $1.07$ points, indicating that the gate is not critically sensitive
to this hyperparameter in this range.
At $\tau = 0.7$, a modest but consistent drop is observed across all orders, suggesting that
too large a $\tau$ reduces the gate's sensitivity to drift; the exponential factor
$\exp(-\delta_i / \tau)$ flattens toward $1$ for all rank directions, causing the gate to
over-retain directions that the new task has substantially overwritten.
We fix $\tau = 0.5$ as the default, as it achieves the best balance between drift sensitivity
and stability, though the method remains competitive across the full range tested.

\begin{table*}[t]
\centering
\caption{(\%) Performance comparison on three CL benchmark datasets using  LLaMA-3.2-3B-Instruct, across different task order,  where $O_i$ represents $i^{th}$ task order.}
\small
\resizebox{\textwidth}{!}{%
\begin{tabular}{l|cccc|cccc|ccc}
\toprule
 & \multicolumn{4}{c|}{Standard CL Benchmark} 
 & \multicolumn{4}{c|}{Large Number of Tasks} 
 & \multicolumn{3}{c}{SuperNI Benchmark} \\
\midrule
Method 
& O1 $\uparrow$ & O2 $\uparrow$ & O3 $\uparrow$ & avg $\uparrow$
& O4 $\uparrow$ & O5 $\uparrow$ & O6 $\uparrow$ & avg $\uparrow$
& O7 $\uparrow$ & O8 $\uparrow$ & avg $\uparrow$\\
\midrule
SeqLoRA    & 61.21& 61.31& 61.50& 61.34& 49.30& 50.10& 49.10& 49.50& 31.00& 31.85& 31.43\\
IncLoRA    & 66.23& 66.40 & 66.44& 66.35& 53.41& 53.10 & 53.15& 53.22& 32.64& 32.75& 32.70\\
O-LoRA     & 68.08& 68.01 & 68.16& 68.03& 56.14& 56.25& 56.67& 56.35& 33.07& 33.51& 33.29\\
Gain-LoRA& 68.45& 68.80& 68.70& 68.65& 53.70& 53.20& 54.10& 53.67& 33.54& 33.70& 33.62\\
CorDA& 68.08& 68.05 & 68.10& 68.07& 54.53& 55.20& 55.16& 55.3& 28.00& 29.02& 28.51\\
 SLAO& 82.12 & 81.48& 83.93 & 82.51& 57.31& 62.13& 59.42& 59.62& 46.07& 45.52&45.80\\
\midrule
KnOTS      & 59.50& 61.50& 59.38& 60.13& 47.07& 46.86& 47.21& 47.04& 18.93& 19.28& 19.11\\
LoRA-LEGO  & 80.48 & 80.40& 80.35& 80.41& 54.37& 54.63& 54.17& 54.39& 39.54& 39.20& 39.37\\
\midrule
SOAR (ours)& \textbf{85.88}&\textbf{86.45}&\textbf{87.13}& 
           \textbf{86.48}& \textbf{59.71}&\textbf{63.31}&\textbf{65.93}& 
           \textbf{62.98}& \textbf{47.04}&\textbf{48.87}& \textbf{47.95}\\
\bottomrule
\end{tabular}}

\label{tab:llama3.2_main_tab}
\end{table*}

\begin{table*}[t]
\centering
\caption{(\%) Performance comparison on three CL benchmark datasets using  LLaMA-3.1-8B-Instruct, across different task order, where  $O_i$ represents $i^{th}$ task order.}
\small
\resizebox{\textwidth}{!}{%
\begin{tabular}{l|cccc|cccc|ccc}
\toprule
 & \multicolumn{4}{c|}{Standard CL Benchmark} 
 & \multicolumn{4}{c|}{Large Number of Tasks} 
 & \multicolumn{3}{c}{SuperNI Benchmark} \\
\midrule
Method 
& O1 $\uparrow$ & O2 $\uparrow$ & O3 $\uparrow$ & avg $\uparrow$
& O4 $\uparrow$ & O5 $\uparrow$ & O6 $\uparrow$ & avg $\uparrow$
& O7 $\uparrow$ & O8 $\uparrow$ & avg $\uparrow$\\
\midrule
SeqLoRA    & 77.3& 76.75& 77.9& 77.32& 61.38& 60.4& 61.76& 61.18& 29.43& 30.3& 29.87\\
IncLoRA    & 78.34& 78.11& 77.8& 78.08& 63.45& 64.11& 63.76& 63.77& 31.34& 32.44& 31.89\\
O-LoRA     & 79.82& 80.11& 79.21& 79.71& 66.28& 65.16& 65.67& 65.70& 34.83& 36.11& 35.47\\
Gain-LoRA& 80.10& 80.55& 80.40& 80.35& 65.47& 66.24& 65.79& 65.83& 35.15& 36.23& 35.69\\
CorDA& 79.87& 79.98& 79.82& 79.89& 65.08& 64.65& 65.67& 65.13& 31.98& 31.05& 31.52\\
 SLAO& 87.08& 87.53& 87.35& 87.32& 65.25& 68.76& 65.8& 66.60& 47.50& 48.18&47.84\\
\midrule
KnOTS      & 77.62& 77.50& 77.75& 77.62& 60.42& 60.49& 60.69& 60.54& 28.64& 28.61& 28.63\\
LoRA-LEGO  & 85.52& 85.32& 85.4& 85.41& 63.37& 63.28& 63.46& 63.37& 41.90& 42.24& 42.07\\
\midrule
SOAR (ours)& \textbf{89.35}&\textbf{90.10}&\textbf{89.70}& 
           \textbf{89.72}& \textbf{68.42}&\textbf{69.88}&\textbf{68.49}& 
           \textbf{68.93}& \textbf{49.98}&\textbf{51.83}& \textbf{50.90}\\
\bottomrule
\end{tabular}}

\label{tab:llama3.1_main_tab}
\end{table*}

\begin{figure}[h]
    \centering
    \includegraphics[width=\linewidth]{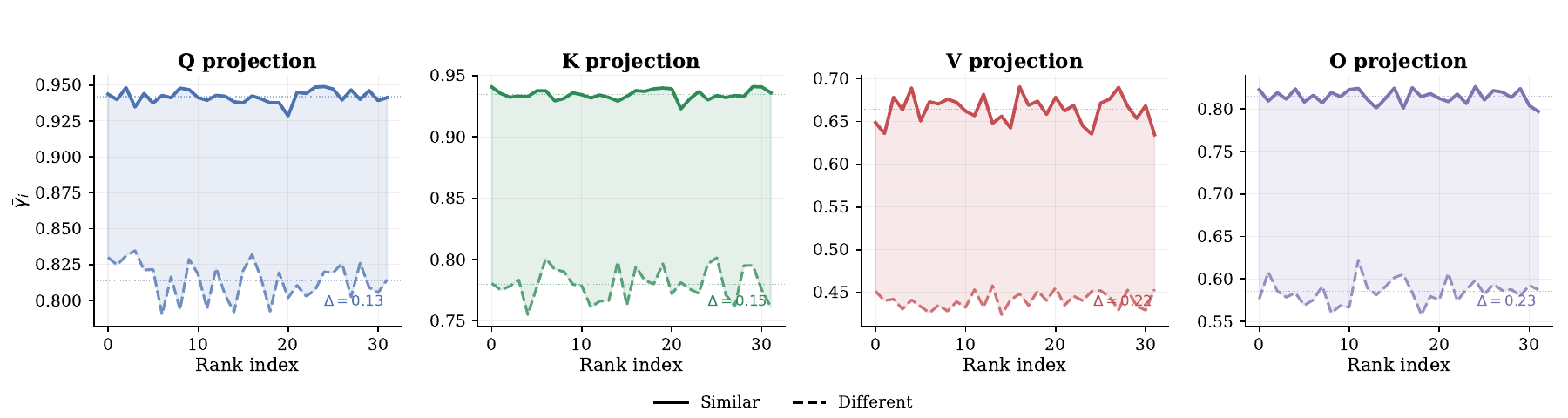}
    \caption{$\gamma$ variance on similar vs dissimilar task, averaged across layers for rank 32}
    \label{fig:g_ablation}
\end{figure}

\subsection{Detailed per-rank gating analysis: }
\label{app:detail_gamma}
Figure~\ref{fig:g_ablation} reports the rank-wise retention gate
$\gamma_i = \max(c_i, 0) \cdot \exp(-\delta_i / \tau)$, averaged across all transformer layers,
for all four projection matrices (\texttt{Q\_proj}, \texttt{K\_proj}, \texttt{V\_proj},
\texttt{O\_proj}) under same-task and different-task transition regimes. Two consistent patterns emerge across all projection types.
First, the same-task $\gamma_i$ (blue) is uniformly and substantially higher than the different-task
$\gamma_i$ (orange) across all 32 rank directions, confirming that the gate correctly
discriminates task-relatedness: when consecutive tasks are similar, the aligned old directions
$\widetilde{\mathbf{b}}_i^{\mathrm{old}}$ exhibit high cosine agreement $c_i$ with
$\mathbf{b}_i^{\mathrm{ft}}$ and small forgetting score $\delta_i$, driving $\gamma_i$ toward
$1$ and encouraging strong retention of prior knowledge.
Conversely, under dissimilar-task transitions, $\gamma_{i}$ is lower, which means even after Procrustes alignment, the old and new factors still differ significantly at the local scales, implying the need for local adjustment for proper merging. 
Second, the absolute level of $\gamma_i$ varies meaningfully across projection types, revealing a structural hierarchy in how much prior information is retained.
\texttt{Q\_proj} and \texttt{K\_proj} attain the highest same-task gates
($\gamma_i \approx 0.94$ and $0.93$ respectively), suggesting that the query and key subspaces are the most geometrically stable across tasks and thus accumulate the most reusable structure.
\texttt{O\_proj} occupies an intermediate regime ($\gamma_i \approx 0.81$), while
\texttt{V\_proj} exhibits the lowest gates overall ($\gamma_i \approx 0.66$), indicating
that value projections undergo the most task-specific rewriting and are therefore
appropriately down-weighted during consolidation.
Critically, this projection-wise differentiation is not manually specified but emerges automatically from the gate's construction, validating that the Procrustes-aligned coordinate frame provides a consistent basis in which $c_i$ and $\delta_i$ reflect genuine task-specific geometry rather than coordinate artifacts. Here, two Similar tasks considered are (amazon polarity, imdb movie reviews); dissimilar tasks consist of (xsum summarization, emotion classification).

\begin{algorithm}[H]
\caption{SOAR: Stiefel Optimization and Aligned Rotation for Continual LoRA Learning}
\label{alg:soar}
\begin{algorithmic}[1]

\Require Pre-trained weights $\mathbf{W}_0$, tasks $\{\mathcal{D}_t\}_{t=1}^{T}$, rank $r$, lr $\eta$, temperature $\tau$
\Ensure $\mathbf{A}_{\mathrm{merge}}^{(T)} \in \mathrm{St}(n,r)$,\; $\mathbf{B}_{\mathrm{merge}}^{(T)} \in \mathbb{R}^{m \times r}$

\State Init $\mathbf{A} \sim \mathrm{St}(n,r)$,\; $\mathbf{B} \leftarrow \mathbf{0}$,\; $\mathbf{A}_{\mathrm{merge}} \leftarrow \varnothing$,\; $\mathbf{B}_{\mathrm{merge}} \leftarrow \varnothing$

\For{$t = 1, \ldots, T$}

    \State $(\mathbf{A},\, \mathbf{B}) \leftarrow \bigl(\mathbf{A}_{\mathrm{merge}}^{(t-1)},\, \mathbf{B}_{\mathrm{merge}}^{(t-1)}\bigr)$ \textbf{if} $t > 1$,\; \textbf{else} $\bigl(\mathbf{A}^{(0)},\, \mathbf{0}\bigr)$

    \For{each mini-batch $\mathcal{B} \sim \mathcal{D}_t$}
        \Comment{Riemannian Adam}
        \State Update $\mathbf{B}$ via Euclidean Adam on $\nabla_{\!\mathbf{B}}\,\mathcal{L}_t$
        \State Update $\mathbf{A}$ via Riemannian gradient $+$ polar retraction onto $\mathrm{St}(n,r)$
    \EndFor

    \State $\mathbf{A}_{\mathrm{merge}}^{(t)} \leftarrow \mathbf{A}_{\mathrm{ft}}^{(t)}$
    \Comment{Direct replacement}

    \If{$t = 1$}
        \State $\mathbf{B}_{\mathrm{merge}}^{(1)} \leftarrow \mathbf{B}_{\mathrm{ft}}^{(1)}$

    \Else
        \Comment{Procrustes-aligned adaptive merge}

        \State \textbf{SVD:}\; $\bigl(\mathbf{B}_{\mathrm{ft}}^{(t)}\bigr)^{\!\top} \mathbf{B}_{\mathrm{merge}}^{(t-1)} = \mathbf{U}\boldsymbol{\Sigma}\mathbf{V}^{\top}$,\quad $\mathbf{R}^{*} = \mathbf{U}\mathbf{V}^{\top}$
        \Comment{Procrustes on $\mathbf{B}$}

        \State $\widetilde{\mathbf{B}}_{\mathrm{merge}}^{(t-1)} \leftarrow \mathbf{B}_{\mathrm{merge}}^{(t-1)}\,(\mathbf{R}^{*})^{\top}$
        \Comment{Align old $\mathbf{B}$ to new frame}


        \State $c_i \leftarrow \bigl({\mathbf{b}}^{\mathrm{ft}}_i\bigr)^{\!\top}{\tilde{\mathbf{b}}}^{\mathrm{old}}_i \;\in [-1, 1]$\quad for each $i$
        \Comment{Per-rank cosine similarity}

        \State $\delta_i \leftarrow \dfrac{\|\mathbf{b}^{\mathrm{ft}}_i - \tilde{\mathbf{b}}^{\mathrm{old}}_i\|^2}{\|\tilde{\mathbf{b}}^{\mathrm{old}}_i\|^2 + \epsilon}$\quad for each $i$
        \Comment{Per-rank forgetting score}

        \State $\gamma_i \leftarrow \max(c_i,\, 0) \cdot \exp\!\left(-\delta_i\, /\, \tau\right) \;\in [0, 1]$\quad for each $i$
        \Comment{Retention gate}

        \State $\mathbf{B}_{\mathrm{merge}}^{(t)} \leftarrow \mathbf{B}_{\mathrm{ft}}^{(t)} + \bigl(\widetilde{\mathbf{B}}_{\mathrm{merge}}^{(t-1)} - \mathbf{B}_{\mathrm{ft}}^{(t)}\bigr)\,\mathrm{diag}(\boldsymbol{\gamma})$
        \Comment{Adaptive additive correction}

    \EndIf

\EndFor

\State \Return $\mathbf{A}_{\mathrm{merge}}^{(T)}$,\; $\mathbf{B}_{\mathrm{merge}}^{(T)}$

\end{algorithmic}
\end{algorithm}

\begin{figure}[h]
    \centering
    \includegraphics[width=\linewidth]{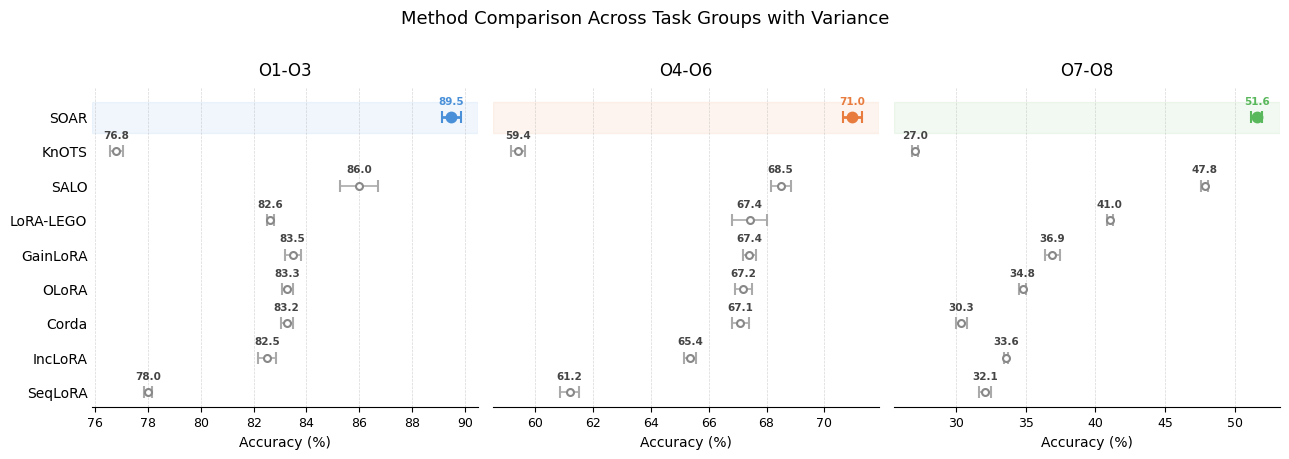}
    \caption{(\%) Performance analysis averaged across the benchmark dataset.}
    \label{fig:var_main_fig}
\end{figure}

\begin{figure}[h]
    \centering
    \includegraphics[width=0.8\linewidth]{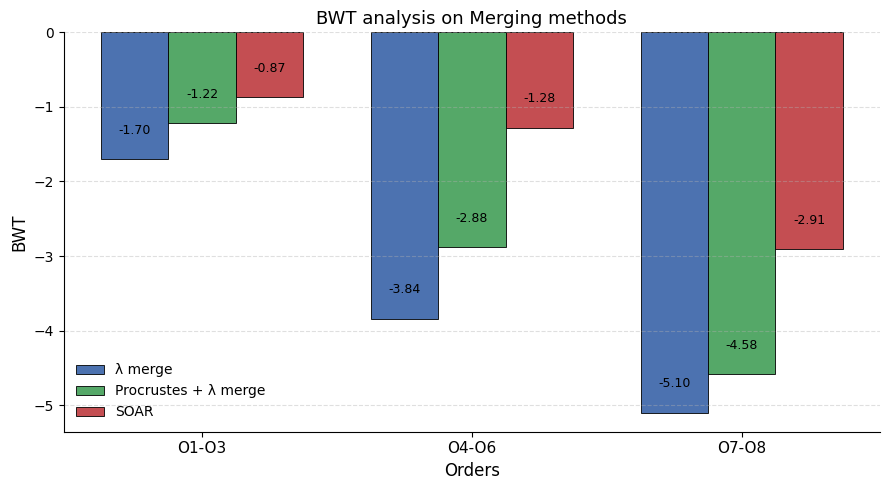}
    \caption{Effect of Procrustes and Per Rank Gating using Qwen3-8B across varying datasets.}
    \label{fig:b_ablation}
\end{figure}

\begin{table}[t]
\centering
\caption{Descriptions of datasets used from standard continual learning benchmarks, GLUE, and SuperGLUE.}
\resizebox{\textwidth}{!}{%
\begin{tabular}{l l l l l}
\toprule
\textbf{Dataset name} & \textbf{Category} & \textbf{Task} & \textbf{Domain} & \textbf{Metric} \\
\midrule
Amazon Polarity & CL Benchmark & Sentiment analysis & Amazon reviews & Accuracy \\
DBpedia & CL Benchmark & Topic classification & Wikipedia & Accuracy \\
Yahoo Answers & CL Benchmark & Topic classification & Yahoo Q\&A & Accuracy \\
AG News & CL Benchmark & Topic classification & News & Accuracy \\

MNLI & GLUE & Natural language inference & Various & Accuracy \\
QQP & GLUE & Paraphrase detection & Quora & Accuracy \\
RTE & GLUE & Natural language inference & News, Wikipedia & Accuracy \\
SST-2 & GLUE & Sentiment analysis & Movie reviews & Accuracy \\

WiC & SuperGLUE & Word sense disambiguation & Lexical databases & Accuracy \\
BoolQ & SuperGLUE & Boolean question answering & Wikipedia & Accuracy \\
MultiRC & SuperGLUE & Question answering & Various & Accuracy \\
ReCoRD & SuperGLUE & Reading comprehension & News & Accuracy \\

IMDB & Sentiment & Sentiment analysis & Movie reviews & Accuracy \\
\bottomrule
\end{tabular}
}
\end{table}

\subsection{Compute Overhead Analysis: }
\label{app:over_head}

Table~\ref{tab:compute_total} reports total continual learning runtime for each method, measured on the O1 task order with 4 tasks, batch size 2, and 1{,}250 training steps per task on a single H100 GPU, using Qwen3-8B under an identical experimental configuration. All methods fall within a narrow band of roughly $1{,}200$--$1{,}600$s, confirming that the choice of consolidation strategy has a second-order effect on cost relative to the dominant $O(bnm)$ forward and backward passes. SOAR is the most expensive at $1{,}582$s, a $15.6\%$ overhead over vanilla sequential fine-tuning ($1{,}368$s) and between $7.5\%$ and $30\%$ over the other baselines. This overhead is fully accounted for by two operations: the Riemannian gradient projection and polar retraction applied to $\mathbf{A}$ at each step, and a single per-task Procrustes SVD on an $(r \times r)$ matrix. Both scale as $O(r^3)$ in the rank and are independent of sequence length and batch size, so the relative cost decreases as model size or token budget grows. The per-step manifold operations account for the larger share, as they recur at every optimizer step, whereas the Procrustes SVD is incurred once per task boundary and is negligible at $r{=}32$. Methods requiring additional per-step machinery are comparable: O-LoRA ($1{,}452$s) pays for its orthogonality penalty across all previous adapters, and LoRA-LEGO ($1{,}472$s) for rank-wise clustering at each merge. Methods with lighter-weight consolidation --- IncLoRA ($1{,}219$s), SLAO ($1{,}254$s), GainLoRA ($1{,}299.5$s) and KnOTS ($1{,}300$s) --- are marginally cheaper, with IncLoRA's savings offset by adapter memory that grows linearly in the number of tasks. SOAR's overhead is therefore modest, bounded, and predictable: it is fixed by the rank rather than the data, and does not accumulate with sequence length.

\begin{table}[h]
\centering
\caption{Total compute time taken}
\begin{tabular}{lc}
\toprule
\textbf{Method} & \textbf{Time Taken (seconds)}\\
\midrule
KnoTS & 1300.0\\
LoRA-LEGO & 1472.0\\
GainLoRA & 1299.5\\
IncLoRA & 1219.0\\
SLAO & 1254.0\\
OLoRA & 1452.0\\
SOAR (ours) & 1582.0\\
\bottomrule
\end{tabular}

\label{tab:compute_total}
\end{table}



\subsection{Backward transfer analysis}
\label{app:bwt_comp}
Table~\ref{tab:bwt_comparison} reports Backward Transfer (BWT) analysis for all three LLM families on all task orders. Here, \textbf{A} is Qwen3-8B, \textbf{B} is Llama-3.1-8B, and \textbf{C} is Llama-3.2-3B, SCL is the standard continual learning task; LNT is large number of tasks; and SNI is the SuperNI benchmark dataset, where less negative values
indicate better retention of prior knowledge. Three patterns hold
consistently. First, the ordering of methods is stable across all three
backbones and all nine orders: SeqLoRA suffers the most severe
forgetting, confirming that unconstrained sequential fine-tuning of a
single adapter catastrophically overwrites prior tasks; memory-isolation
methods such as IncLoRA and CorDA improve on it but still forget
substantially, as they do not explicitly consolidate knowledge across
tasks; O-LoRA and Gain-LoRA reduce forgetting further through
orthogonality constraints and gating; and SLAO is the strongest baseline
throughout, benefiting from continual merging within a single adapter.
The offline merging methods behave as expected in this sequential,
data-free setting: KnOTS degrades sharply, while LoRA-LEGO remains
competitive with the continual-learning baselines. Second, forgetting
grows with benchmark difficulty for every method, from the Standard CL
benchmark through the Large Number of Tasks setting to SuperNI, where
longer and more heterogeneous task sequences leave the most room for
interference. Third, SOAR attains the lowest forgetting in every column,
reducing BWT relative to SLAO by $52$--$70\%$ on the Standard CL and
Large Number of Tasks benchmarks and by $31$--$43\%$ on SuperNI, and by
nearly an order of magnitude relative to SeqLoRA. Crucially, SOAR's
margin is preserved across model families and scales, indicating that
Procrustes-aligned, rank-wise gated merging provides more faithful
knowledge consolidation than either orthogonality-based or uniform
merging, independently of the backbone.



\subsection{Ablation for update rule}
\label{app:update_rule_ablations}
To directly examine the assumption underlying our update rule, we first measure the principal-angle drift between the input subspaces spanned by consecutive $\mathbf{A}$ matrices. Principal angles isolate genuine subspace drift while remaining invariant to rotations within the subspace, making them an appropriate measure for this question. As shown in Table \ref{tab:A_angle_drift}, the measured drift is small on O1 and O4, where tasks within each sequence are relatively similar, indicating that the learned input subspace is largely preserved across tasks. In contrast, O7 exhibits noticeably larger drift, consistent with its substantially greater task diversity.

To further validate this observation, in Table \ref{tab:A_initialisation} we evaluate three alternative update strategies for $\mathbf{A}$. A fixed random orthonormal $\mathbf{A}$ performs the worst, demonstrating that a randomly chosen input subspace is far from the task-specific subspace learned during adaptation. Freezing $\mathbf{A}$ after the first task results in only a modest degradation O1 and O4, where the measured subspace drift is limited, but degrades much more noticeably on O7. This behavior supports the intuition that while previously learned input directions remain useful for related tasks, adapting $\mathbf{A}$ becomes increasingly important as the task distribution evolves and new input subspaces emerge.

We also investigate explicit merging of $\mathbf{A}$ using the same Procrustes alignment framework employed for $\mathbf{B}$. We observe improvements over direct replacement, with relatively modest gains on O1 and O4 and larger gains on O7, mirroring the measured increase in subspace drift. These results suggest that explicit $\mathbf{A}$ merging becomes more beneficial as tasks become more heterogeneous. However, this improvement comes at the cost of an additional interpolation hyperparameter and an extra Procrustes alignment and Stiefel retraction step, increasing the runtime by approximately 134.6 seconds (Qwen 3 8B for task order O1) per continual learning run. Our method is motivated by the asymmetry between the two LoRA factors \citep{qiao2025mergeforgetsinglelora}, which suggests that the input factor $\mathbf{A}$ remains relatively stable across tasks. Our empirical analysis is consistent with this observation: as shown in Table \ref{tab:A_angle_drift}, the measured principal-angle drift of $\mathbf{A}$ is small on O1 and O4, while a larger, though still moderate, drift is observed on the more heterogeneous O7 benchmark.

\begin{table}[h]
\centering
\caption{Ablations on initialisation strategies for $\mathbf{A}$}
\begin{tabular}{lcccc}
\toprule
$\mathbf{A}$ \textbf{update rule} & \textbf{O1} & \textbf{O4} & \textbf{O7} & \textbf{Average}\\
\midrule
Random fixed $\mathbf{A}$ & 70.80 & 58.43 & 34.56 & 54.59
\\
Freeze after first task	 & 87.45 & 67.86 & 42.91 & 65.92\\
SOAR & 89.35 & 71.23 & 51.28 & 70.62\\
SOAR + Procrustes $\mathbf{A}$-merge & \textbf{89.90} & \textbf{72.80} & \textbf{53.10} & \textbf{71.93}\\
\bottomrule
\end{tabular}

\label{tab:A_initialisation}
\end{table}

\begin{table}[h]
\centering
\caption{Measured principal-angle drift of $\mathbf{A}$}
\begin{tabular}{lccc}
\toprule
\textbf{Metric} & \textbf{O1} & \textbf{O4} & \textbf{O7}\\
\midrule
Mean principal-angle drift of $\mathbf{A}$ & $3.50^\circ$ & $5.04^\circ$ & $12.59^\circ$\\
\bottomrule
\end{tabular}

\label{tab:A_angle_drift}
\end{table}

\begin{table}[t]
\centering
\caption{Descriptions of datasets used from SNLI-derived SuperNI benchmark.}
\resizebox{\textwidth}{!}{%
\begin{tabular}{l l l l}
\toprule
\textbf{Dataset name} & \textbf{Task} & \textbf{Category} & \textbf{Metric} \\
\midrule
multi-woz-user-utterance-generation & Dialogue generation & SuperNI & Rouge-L \\
diplomacy-text-generation & Dialogue generation & SuperNI & Rouge-L \\
personachat-generate-next & Dialogue generation & SuperNI & Rouge-L \\

outcome-extraction & Information extraction & SuperNI & Rouge-L \\
glucose-reverse-cause-event-detection & Information extraction & SuperNI & Rouge-L \\
evaluation-relation-extraction & Information extraction & SuperNI & Rouge-L \\

quoref-answer-generation & Question answering & SuperNI & Rouge-L \\
commonsenseqa-answer-generation & Question answering & SuperNI & Rouge-L \\
sciq-answer-generation & Question answering & SuperNI & Rouge-L \\

reddit-tifu-long-text-summarization & Summarization & SuperNI & Rouge-L \\
xsum-summarization & Summarization & SuperNI & Rouge-L \\
samsum-summary & Summarization & SuperNI & Rouge-L \\

sst2-polarity-classification & Sentiment analysis & SuperNI & Accuracy \\
emotion-classification & Sentiment analysis & SuperNI & Accuracy \\
\bottomrule
\end{tabular}
}
\end{table}

\begin{table}[H]
\centering
\caption{Eight different task orders used in continual learning experiments.}
\label{tab:orders}
\begin{tabular}{c p{0.8\textwidth}}
\toprule
\textbf{Order} & \textbf{Task Sequence} \\
\midrule

1 & dbpedia $\rightarrow$ ag $\rightarrow$ amazon $\rightarrow$ yahoo \\

2 & dbpedia $\rightarrow$ amazon $\rightarrow$ ag $\rightarrow$ yahoo \\

3 & ag $\rightarrow$ yahoo $\rightarrow$ dbpedia $\rightarrow$ amazon \\

\midrule

4 & amazon $\rightarrow$ qqp $\rightarrow$ mnli $\rightarrow$ imdb $\rightarrow$ sst-2 $\rightarrow$ wic $\rightarrow$ yahoo $\rightarrow$ boolq \\
  & $\rightarrow$ ag $\rightarrow$ dbpedia $\rightarrow$ rte $\rightarrow$ multirc $\rightarrow$ record \\

5 & mnli $\rightarrow$ imdb $\rightarrow$ rte $\rightarrow$ record $\rightarrow$ boolq $\rightarrow$ qqp $\rightarrow$ sst-2 $\rightarrow$ yahoo \\
  & $\rightarrow$ multirc $\rightarrow$ ag $\rightarrow$ amazon $\rightarrow$ dbpedia $\rightarrow$ wic \\

6 & multirc $\rightarrow$ wic $\rightarrow$ yahoo $\rightarrow$ qqp $\rightarrow$ rte $\rightarrow$ ag $\rightarrow$ dbpedia $\rightarrow$ amazon \\
  & $\rightarrow$ mnli $\rightarrow$ record $\rightarrow$ imdb $\rightarrow$ boolq $\rightarrow$ sst-2 \\

\midrule

7 (SuperNI) & sciq $\rightarrow$ commonsenseqa $\rightarrow$ quoref $\rightarrow$ samsum $\rightarrow$ relation $\rightarrow$ reddit $\rightarrow$ xsum \\
            & $\rightarrow$ diplomacy $\rightarrow$ multiwoz $\rightarrow$ emotion $\rightarrow$ personachat $\rightarrow$ outcome $\rightarrow$ glucose $\rightarrow$ sst2 \\

8 (SuperNI) & quoref $\rightarrow$ samsum $\rightarrow$ multiwoz $\rightarrow$ xsum $\rightarrow$ outcome $\rightarrow$ diplomacy \\
            & $\rightarrow$ commonsenseqa $\rightarrow$ relation $\rightarrow$ glucose $\rightarrow$ personachat $\rightarrow$ emotion $\rightarrow$ sciq $\rightarrow$ reddit $\rightarrow$ sst2 \\

\bottomrule
\end{tabular}
\end{table}

\section{Limitations}
While SOAR provides an efficient and principled framework for continual LoRA consolidation, it has several limitations. First, the method operates at the level of individual rank directions, implicitly assuming that useful information can be localized within these components. In practice, however, task-relevant information may be distributed across multiple rank directions, forming shared subspaces rather than axis-aligned components. As a result, per-direction gating may not fully capture cross-directional dependencies, potentially leading to suboptimal retention of distributed knowledge. 

Second, SOAR does not rely on replay or stored data, and therefore cannot recover information that has been fully overwritten during adaptation. Once a direction is reassigned to a new task and its prior contribution is lost, this information cannot be reconstructed in subsequent steps. Addressing these limitations, for example, through subspace-level consolidation or lightweight memory mechanisms, is an important direction for future work.

\section{Licenses for Existing Assets}
\label{app:licenses}

All assets used in this work are properly credited with citations to their
original papers. We list the primary assets and their licenses below.

\textbf{Datasets and Evaluation dataset}
(1) DBpedia, AG News, Amazon Polarity, Yahoo Answers. \citep{zhang2016characterlevelconvolutionalnetworkstext} (2) \textbf{Large number of tasks: } It contains 4 standard CL tasks, along with GLUE tasks \citep{wang2019glue} (MNLI, QQP, SST-2, RTE) (CC BY-NC 4.0) and 4 SuperGLUE tasks \citep{wang2019superglue} (WiC, ReCoRD, MultiRC, BoolQ) and IMDB movie reviews (CC BY-NC 4.0). Each of these tasks uses 2500 randomly sampled training samples and 500 test samples per class (3) \textbf{SuperNI Benchmark} \citep{wang2022supernaturalinstructionsgeneralizationdeclarativeinstructions} (Apache-2.0 license).

\textbf{Models.}
Qwen3-8B~\citep{yang2025qwen3technicalreport} is released under Apache 2.0
(\url{https://huggingface.co/Qwen/Qwen3-4B}).
LLaMA-3.2-3B~\citep{grattafiori2024llama3herdmodels} is released under the
Llama 3.2 Community License
(\url{https://huggingface.co/meta-llama/Llama-3.2-3B}).
Llama-3.1-8B~\citep{grattafiori2024llama3herdmodels} is released under the
Llama 3.1 Community License Agreement (\url{https://huggingface.co/meta-llama/Llama-3.1-8B-Instruct})

\textbf{Baselines.}
(1) Continual learning baselines: SLAO \citep{qiao2025mergeforgetsinglelora} (Since the original implementation is not publicly available, we implemented the method ourselves for our evaluation.), SeqLoRA; IncLoRA (are common and relatively simple baselines, so we implemented them ourselves for our experiments).

O-LoRA \citep{wang2023orthogonalsubspacelearninglanguage} (\url{https://github.com/cmnfriend/O-LoRA}) with MIT license; 
GainLoRA \citep{liang2025gatedintegrationlowrankadaptation} (\url{https://github.com/liangyanshuo/gainlora}) with MIT license; 
CorDA \citep{yang2025cordacontextorienteddecompositionadaptation} (\url{https://github.com/iboing/CorDA}) with Apache-2.0 license; 

(2)LoRA merging baselines: LoRA-LEGO \citep{zhao2024mergingloraslikeplaying} (codes are not publicly available; hence it is coded by us); 
 KnOTS \citep{stoica2024modelmergingsvdtie} (\url{https://github.com/gstoica27/knots}) with MIT license.

All model and dataset versions correspond to those publicly available at the time of submission.




\newpage
\section*{NeurIPS Paper Checklist}

\begin{enumerate}

\item {\bf Claims}
    \item[] Question: Do the main claims made in the abstract and introduction accurately reflect the paper's contributions and scope?
    \item[] Answer: \answerYes{} 
    \item[] Justification: We have identified a problem of non-identifiably in LoRA based continual merging which leads to ill-posed merging as well as catastrophic forgetting. We have tried to theoretically explain via Lemma\ref{lemma:keylemma} and via extensive experiments in Section 5.

\item {\bf Limitations}
    \item[] Question: Does the paper discuss the limitations of the work performed by the authors?
    \item[] Answer: \answerYes{} 
    \item[] Justification: Limitations have briefly discussed in Conclusion and more detailes are provided in Appendix section tilted "Limitations".

\item {\bf Theory assumptions and proofs}
    \item[] Question: For each theoretical result, does the paper provide the full set of assumptions and a complete (and correct) proof?
    \item[] Answer: \answerYes 
    \item[] Justification: We have provided proof of Lemma\ref{lemma:keylemma} and justified other claims empirically also.

    \item {\bf Experimental result reproducibility}
    \item[] Question: Does the paper fully disclose all the information needed to reproduce the main experimental results of the paper to the extent that it affects the main claims and/or conclusions of the paper (regardless of whether the code and data are provided or not)?
    \item[] Answer: \answerYes{} 
    \item[] Justification: See section 5.

\item {\bf Open access to data and code}
    \item[] Question: Does the paper provide open access to the data and code, with sufficient instructions to faithfully reproduce the main experimental results, as described in supplemental material?
    \item[] Answer: \answerNo{} 
    \item[] Justification: We will subject to internal approval.

\item {\bf Experimental setting/details}
    \item[] Question: Does the paper specify all the training and test details (e.g., data splits, hyperparameters, how they were chosen, type of optimizer) necessary to understand the results?
    \item[] Answer: \answerYes{} 
    \item[] Justification: See Section 5 and Appendix.

\item {\bf Experiment statistical significance}
    \item[] Question: Does the paper report error bars suitably and correctly defined or other appropriate information about the statistical significance of the experiments?
    \item[] Answer: \answerYes{} 
    \item[] Justification: We have ran the experiments on multiple seeds and reported the mean and standard deviations for each method.

\item {\bf Experiments compute resources}
    \item[] Question: For each experiment, does the paper provide sufficient information on the computer resources (type of compute workers, memory, time of execution) needed to reproduce the experiments?
    \item[] Answer: \answerYes{} 
    \item[] Justification: All experiment were run on 1 H100 GPU of 80GB. Details are provided in Section 5.
    
\item {\bf Code of ethics}
    \item[] Question: Does the research conducted in the paper conform, in every respect, with the NeurIPS Code of Ethics \url{https://neurips.cc/public/EthicsGuidelines}?
    \item[] Answer: \answerNA{} 
    \item[] Justification: The paper does not involve human participants, all data used are open source popular datasets for Continual Learning tasks.

\item {\bf Broader impacts}
    \item[] Question: Does the paper discuss both potential positive societal impacts and negative societal impacts of the work performed?
    \item[] Answer: \answerNA{} 
    \item[] Justification: This work does not deal with safety issues, discrimination, bias etc to the best of our knowledge. 
    
\item {\bf Safeguards}
    \item[] Question: Does the paper describe safeguards that have been put in place for responsible release of data or models that have a high risk for misuse (e.g., pre-trained language models, image generators, or scraped datasets)?
    \item[] Answer: \answerNA{} 
    \item[] Justification: We use general models for our experiments and have cited all models and data appropriately.

\item {\bf Licenses for existing assets}
    \item[] Question: Are the creators or original owners of assets (e.g., code, data, models), used in the paper, properly credited and are the license and terms of use explicitly mentioned and properly respected?
    \item[] Answer: \answerYes{} 
    \item[] Justification: All models, datasets and codes for baselines used are taken from open source and cited.

\item {\bf New assets}
    \item[] Question: Are new assets introduced in the paper well documented and is the documentation provided alongside the assets?
    \item[] Answer: \answerNA{} 
    \item[] Justification: No new asset introduced in paper.

\item {\bf Crowdsourcing and research with human subjects}
    \item[] Question: For crowdsourcing experiments and research with human subjects, does the paper include the full text of instructions given to participants and screenshots, if applicable, as well as details about compensation (if any)? 
    \item[] Answer: \answerNA{} 
    \item[] Justification: The paper does not involve crowdsourcing nor research with human subjects.

\item {\bf Institutional review board (IRB) approvals or equivalent for research with human subjects}
    \item[] Question: Does the paper describe potential risks incurred by study participants, whether such risks were disclosed to the subjects, and whether Institutional Review Board (IRB) approvals (or an equivalent approval/review based on the requirements of your country or institution) were obtained?
    \item[] Answer: \answerNA{} 
    \item[] Justification: The paper does not involve crowdsourcing nor research with human subjects.

\item {\bf Declaration of LLM usage}
    \item[] Question: Does the paper describe the usage of LLMs if it is an important, original, or non-standard component of the core methods in this research? Note that if the LLM is used only for writing, editing, or formatting purposes and does \emph{not} impact the core methodology, scientific rigor, or originality of the research, declaration is not required.
    \item[] Answer: \answerNA{} 
    \item[] Justification: The core method development in this research does not involve LLMs as any important, original, or non-standard components.

\end{enumerate}

\end{document}